\documentclass[10pt]{article} 
\usepackage[preprint]{tmlr}

\usepackage{amsmath,amsfonts,bm}

\def\eqref#1{equation~\ref{#1}}

\def\floor#1{\lfloor #1 \rfloor}
\def\1{\bm{1}}

\DeclareMathAlphabet{\mathsfit}{\encodingdefault}{\sfdefault}{m}{sl}
\SetMathAlphabet{\mathsfit}{bold}{\encodingdefault}{\sfdefault}{bx}{n}

\newcommand{\sigmoid}{\sigma}

\usepackage{hyperref}
\usepackage{url}

\usepackage{booktabs}       
\usepackage{amsfonts}       
\usepackage{nicefrac}       
\usepackage{xcolor}         
\usepackage{graphicx}
\usepackage{subfigure}
\usepackage{amssymb}
\usepackage{footnote}
\usepackage[ruled,vlined]{algorithm2e}
\usepackage{mathrsfs}
\usepackage{hhline}
\usepackage{dblfloatfix}
\usepackage{color, colortbl}
\usepackage{comment}
\usepackage{threeparttable}
\usepackage{rotating}
\usepackage{wrapfig}
\usepackage{float}

\usepackage{amsmath}
\usepackage{amssymb}
\usepackage{mathtools}
\usepackage{amsthm}

\usepackage{graphicx}
\usepackage{caption}

\theoremstyle{plain}
\newtheorem{theorem}{Theorem}[section]

\theoremstyle{definition}

\theoremstyle{remark}
\newtheorem{remark}[theorem]{Remark}

\title{
 ISR: Invertible Symbolic Regression
}

\author{\name Tony Tohme\thanks{Equal contribution.}  \email tohme@mit.edu \\
      \addr Massachusetts Institute of Technology, USA.
      \AND
      \name Mohammad Javad Khojasteh$^*$ \email m1khojasteh@ucsd.edu \\
      \addr Scripps Institution of Oceanography, USA.
      \\
       \addr Massachusetts Institute of Technology, USA.
      \AND
      \name Mohsen Sadr \email mohsen.sadr@psi.ch\\
      \addr Paul Scherrer Institute, Switzerland.
      \AND
      \name Florian Meyer \email flmeyer@ucsd.edu\\
      \addr University of California San Diego, USA.
      \AND
      \name Kamal Youcef-Toumi \email youcef@mit.edu \\
      \addr Massachusetts Institute of Technology, USA.}

\def\month{02}  
\def\year{2024} 

\def\openreview{\url{
https://openreview.net/forum?id=XXXX
}} 

\SetKwInput{KwInput}{Input}                
\SetKwInput{KwOutput}{Output}            

\usepackage[normalem]{ulem}

\usepackage{mathtools}
\allowdisplaybreaks

\begin{document}

\maketitle

\begin{abstract}
We introduce an Invertible Symbolic Regression (ISR) method. It is a machine learning technique that generates analytical relationships between inputs and outputs of a given dataset via invertible maps (or architectures). The proposed ISR method naturally combines the principles of Invertible Neural Networks (INNs) and Equation Learner (EQL), a neural network-based symbolic architecture for function learning. In particular, we transform the affine coupling blocks of INNs into a symbolic framework, resulting in an end-to-end differentiable symbolic invertible architecture that allows for efficient gradient-based learning. The proposed ISR framework also relies on sparsity promoting regularization, allowing the discovery of concise and interpretable invertible expressions. We show that ISR can serve as a (symbolic) normalizing flow for density estimation tasks. Furthermore, we highlight its practical applicability in solving inverse problems, including a benchmark inverse kinematics problem, and notably, a geoacoustic inversion problem in oceanography aimed at inferring posterior distributions of underlying seabed parameters from acoustic signals.
\end{abstract}

\section{Introduction}
\label{sec:introduction}


In many applications in engineering and science, experts have developed theories about how  measurable quantities result from system parameters, known as forward modeling. In contrast, the \emph{Inverse Problem} aims to find unknown parameters of a system that lead to desirable observable quantities.
A typical challenge is that numerous configurations of these parameters yield the same observable quantity, especially with underlying complicated nonlinear governing equations and where hidden parameters outnumber the observable variables. 
To tackle challenging and ill-posed inverse problems, a common method involves estimating a posterior distribution on the unknown parameters, given the observations. Such a probabilistic approach facilitates the uncertainty quantification by analyzing  the diversity of potential inverse solutions.

An established computationally expensive approach in finding the posterior distribution 
is to directly generate samples 
using acceptance/rejection. In this scope, the Markov Chain Monte Carlo (MCMC) methods~\citep{brooks2011handbook,andrieu2003introduction,doucet2005monte,murphy2012machine,goodfellow2016deep,korattikara2014austerity,atchade2005adaptive,kungurtsev2023decentralized}  offer a strong alternative for achieving near-optimum Bayesian estimation~\citep{constantine2016accelerating,conrad2016accelerating,DosDet:11}. However, MCMC methods can be inefficient ~\cite{mackay2003information} as the number of unknown parameter increases.

When the likelihood is unknown or intractable,
the Approximate Bayesian Computation (ABC)~\cite{csillery2010approximate} is often used to estimate the posterior distribution. However, similar to MCMC, this method also suffers from poor scalability~\cite{cranmer2020frontier,papamakarios2019sequential}. A more efficient alternative is to approximate the posterior using a tractable distribution, i.e. the variational method~\cite{blei2017variational,salimans2015markov,wu2018deterministic}. However, the performance of the variational method deteriorates as the true posterior becomes more complicated.

Since direct sampling of the posterior distribution requires many runs of the forward map, often a trained and efficient surrogate model is used instead of the exact model. Surrogate models are considered as an efficient representation of the forward map trained on the data. Popular approaches include recently introduced Physics-Informed Neural Networks \citep{raissi2019physics}. However, due to the black-box nature of neural networks, it is beneficial to express the forward map symbolically instead for several reasons.  First, Symbolic Regression (SR) can provide model interpretability, while understanding the inner workings of deep Neural Networks is challenging~\citep{kim2020integration,gilpin2018explaining}. Second, studying the symbolic outcome can lead to valuable insights and provide nontrivial relations and/or physical laws \citep{udrescu2020aifeynman,udrescu2020aifeynman_2,liu2021machine,keren2023computational}.  Third, they may achieve better results than Neural Networks  in out-of-distribution generalization ~\cite{cranmer2020discovering}. Fourth, unlike conventional regression methods, such as least squares \citep{wild1989nonlinear}, likelihood-based \citep{edwards1984likelihood, pawitan2001all, tohme2023reliable}, and Bayesian regression techniques \citep{lee1997bayesian, leonard2001bayesian,  vanslette2020general,
tohme2020generalized, tohme2020bayesian}, SR does not rely on a fixed parametric model structure. 

The attractive properties of SR, such as interpretability, often come at high computational cost compared to standard Neural Networks. This is because SR optimizes for model structure and parameters simultaneously. Therefore, SR is thought to be NP-hard~\citep{petersen2021deep, virgolin2022symbolic}. However tractable solutions exist, that can approximate the optimal solution suitable for applications. For instance, genetic algorithms \citep{koza1992genetic, schmidt2009distilling, tohme2023gsr,orzechowski2018we,la2021contemporary} and machine learning algorithms, such as neural networks, and transformers \citep{
sahoo2018learning, 
jin2019bayesian,
udrescu2020aifeynman_2, cranmer2020discovering, kommenda2020parameter, operonc++, biggio2021neural, mundhenk2021symbolic, petersen2021deep, valipour2021symbolicgpt,
zhang2022ps, 
kamienny2022end}, are used to solve SR efficiently.

\paragraph{Related Works.} In recent years, a branch of Machine Learning methods has emerged and is dedicated to finding data-driven invertible maps. While they are ideal for data generation and inverse problem, they lack interpretablity. On the other hand, several methods have been developed to achieve interpretablity in representing the forward map via Symbolic Regression. Hence, it is natural to incorporate SR in the invertible map for the inverse problem. Next, we review the related works in the scope of this work.

\textit{Normalizing Flows:} The idea of this class of methods is to train an invertible map such that in the forward problem, the input samples are mapped to a known distribution function, e.g. the normal distribution function. Then, the unknown distribution function is found by inverting the trained map with the normal distribution as the input. This procedure is called  the normalizing flow technique \citep{rezende2015variational,dinh2016density,kingma2018glow,durkan2019neural,tzen2019neural,kobyzev2020normalizing,wang2022minimax}. This method has been used for re-sampling unknown distributions, e.g. Boltzmann generators \citep{noe2019boltzmann}, as well as density recovery such as AI-Feynmann \citep{udrescu2020aifeynman,udrescu2020aifeynman_2}.

\textit{Invertible Neural Networks (INNs):} This method can be categorized in the class of normalizing flows.  The invertibility of INN is rooted in  their architecture. The most popular design is constructed by concatenating  affine coupling
blocks~\cite{kingma2018glow,dinh2016density,dinh2014nice}, which limits the architecture of the neural network. INNs have been shown to be effective in estimating the posterior of the probabilistic inverse problems while outperforming MCMC, ABC, and variational methods. Applications include epidemiology~\cite{radev2021outbreakflow}, astrophysics~\cite{ArdKruRotKot:19}, medicine~\cite{ArdKruRotKot:19}, optics~\cite{Luce_2023}, geophysics~\cite{zhang2021bayesian,WuHuaZha:23}, and reservoir \mbox{engineering~\cite{padmanabha2021solving}}. However, similar to standard neural networks, INNs lead to a model that cannot be evaluated with interpretable mathematical formula.


\textit{Equation Learner (EQL):} Among SR methods, the EQL network is one of the attractive methods since it incorporates gradient descent in the symbolic regression task for better efficiency \citep{martius2016extrapolation, sahoo2018learning, kim2020integration}. EQL devises a neural network-based architecture for SR task by replacing commonly used activation functions with a dictionary of operators and use back-propagation for training. However, in order to obtain a symbolic estimate for the inverse problem efficiently, it is necessary to merge such efficient SR method  with  an invertible architecture, which is the goal of this paper.

\paragraph{Our Contributions.} 
We present Invertible Symbolic Regression (ISR), a machine learning technique that identifies mathematical relationships that best describe the forward and inverse map of a given dataset through the use of invertible maps. ISR is based on an invertible symbolic architecture that bridges the concepts of Invertible Neural Networks (INNs) and Equation Learner (EQL), i.e. a neural network-based symbolic architecture for function learning. In particular, we transform the affine coupling blocks of INNs into a symbolic framework, resulting in an end-to-end differentiable symbolic inverse architecture. This allows for efficient gradient-based learning. The symbolic invertible architecture is easily invertible with a tractable Jacobian, which enables explicit computation of posterior probabilities. The proposed ISR method, equipped with sparsity promoting regularization, captures complex functional relationships with concise and interpretable invertible expressions.
In addition, as a byproduct,  we naturally extend ISR into a conditional ISR (cISR) architecture by integrating the EQL network within conditional INN (cINN) architectures present in the literature.
 We further demonstrate that ISR can also serve as a symbolic normalizing flow (for density estimation) in a number of test distributions.
We demonstrate the applicability of ISR in solving inverse problems, and compare it with INN on a benchmark inverse kinematics problem, as well as a geoacoustic inversion problem in oceanography (see \citep{chapman2021review} for further information). Here, we aim to characterize the undersea environment, such as water-sediment depth, sound speed, etc., from acoustic signals.
To the best of our knowledge, this work is the first attempt towards finding interpretable solutions to general nonlinear inverse problems by establishing analytical relationships between measurable quantities and unknown variables via symbolic invertible maps.

The remainder of the paper is organized as follows. In Section \ref{sec:background}, we go through an overall background about Symbolic Regression (SR) and review the Equation Learner (EQL) network architecture. In Section \ref{sec:isr}, we introduce and present the proposed Invertible Symbolic Regression (ISR) method. We then show our results in Section \ref{sec:results}, where we demonstrate the versatility of ISR as a density estimation method on a variety of examples (distributions), and then show its applicability in inverse problems on an inverse kinematics benchmark problem and through a case study in ocean geoacoustic inversion. Finally in Section \ref{sec:conclusion}, we provide our conclusions and outlook.

\section{Background}
\label{sec:background}
Before diving into the proposed ISR method, we first delve into a comprehensive background of the Symbolic Regression (SR) task, as well as the Equation Learner (EQL) network architecture. 

\paragraph{Symbolic Regression.} Given a dataset $\mathcal{D} = \{\mathbf{x}_i, \mathbf{y}_i\}_{i=1}^N$ consisting of $N$ independent and identically distributed (i.i.d.) paired examples, where $\mathbf{x}_i \in \mathbb{R}^{d_{\mathbf{x}}}$ represents the input variables and $\mathbf{y}_i \in \mathbb{R}^{d_{\mathbf{y}}}$ the corresponding output for the $i$-th observation, the objective of SR is to find an analytical (symbolic) expression $f$ that best maps inputs to outputs, i.e. $\mathbf{y}_i \approx f(\mathbf{x}_i)$. SR seeks to identify the functional form of $f$ from the space of functions $\mathcal{S}$ defined by a set of given arithmetic operations (e.g. $+$, $-$, $\times$, $\div$) and mathematical functions (e.g. $\sin$, $\cos$, $\exp$, etc.) that minimizes a predefined loss function $\mathcal{L}(f, \mathcal{D})$, which measures the discrepancy between the true outputs $\mathbf{y}_i$ and the predictions $f(\mathbf{x}_i)$ over all observations in the dataset. Unlike conventional regression methods that fit parameters within a predefined model structure, SR dynamically constructs the model structure itself, offering a powerful means to uncover underlying physical laws and/or nontrivial relationships. 
\paragraph{Equation Learner Network.}
  \begin{figure}[t]
    \centering    	
    \includegraphics[scale=0.75]{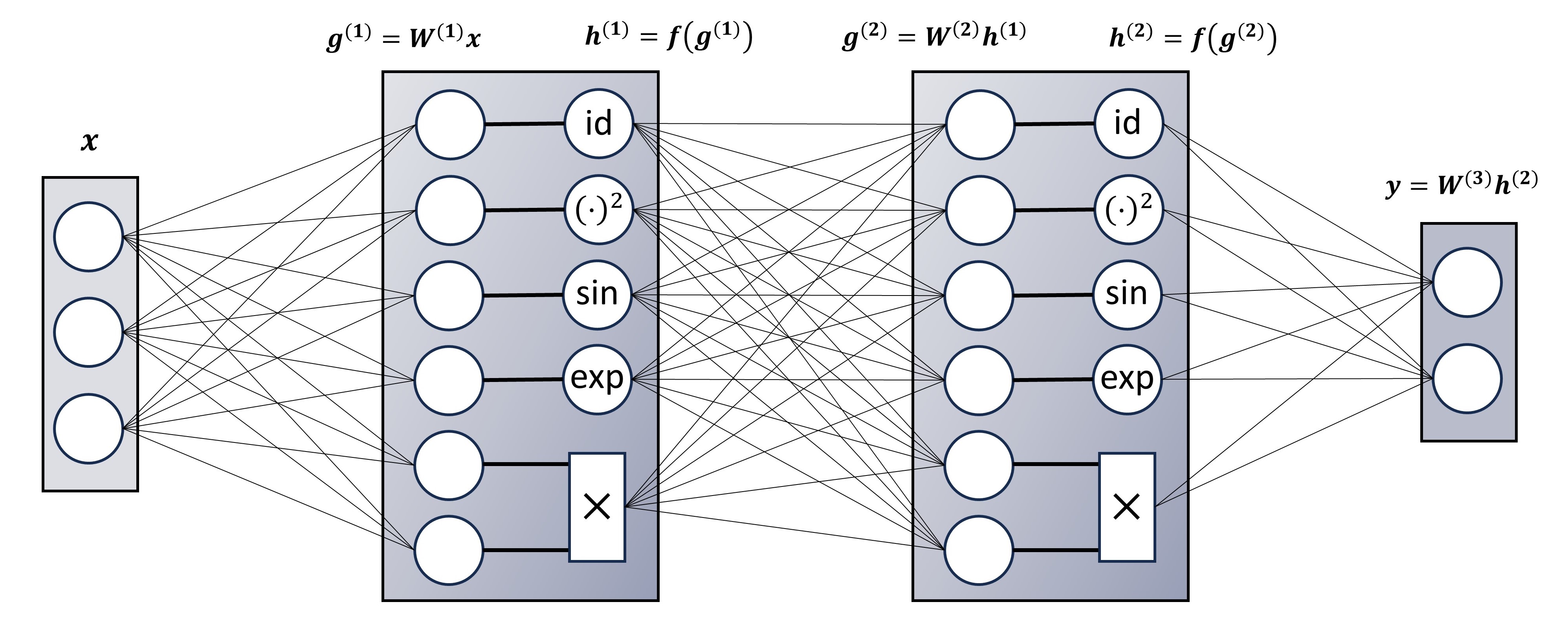}
	\caption{EQL network architecture for symbolic regression. For visual simplicity, we only show 2 hidden layers and 5 activation functions per layer (identity or ``id'', square, sine, exponential, and multiplication).}
 \label{fig:eql_architecture}
\end{figure}

The Equation Learner (EQL) network is a multi-layer feed-forward neural network that is capable of performing symbolic regression by substituting traditional nonlinear activation functions with elementary functions. The EQL network was initially introduced by \citet{martius2016extrapolation} and \citet{sahoo2018learning}, and further explored by \citet{kim2020integration}. As shown in Figure \ref{fig:eql_architecture}, the EQL network architecture is based on a fully connected neural network where the ouput $\mathbf{h}^{(i)}$ of the $i$-th layer is given by
\begin{align}
\mathbf{g}^{(i)} &= \mathbf{W}^{(i)}\mathbf{h}^{(i-1)}\\
\mathbf{h}^{(i)} &= f\scalebox{1.3}{$($}\mathbf{g}^{(i)}\scalebox{1.3}{$)$}
\end{align}
where $\mathbf{W}^{(i)}$ is the weight matrix of the $i$-th layer, $f$ denotes the nonlinear (symbolic) activation functions, and $\mathbf{h}^{(0)} = \mathbf{x}$ represents the input data. In regression tasks, the final layer does not typically have an activation function, so the output for a network with $L$ hidden layers is given by
\begin{align}
\mathbf{y} = \mathbf{h}^{(L+1)} = \mathbf{g}^{(L+1)} = \mathbf{W}^{(L+1)}\mathbf{h}^{(L)}.
\end{align}
In traditional neural networks, activation functions such as ReLU, tanh, or sigmoid are typically employed. However, for the EQL network, the activation function $f(\mathbf{g})$ may consist of a separate primitive function for each component of $\mathbf{g}$ (e.g. the square function, sine, exponential, etc.), and may include functions that take multiple arguments (e.g. the multiplication function). In addition, the primitive functions may be duplicated within each layer (to reduce the training's sensitivity to random initializations).

It is worth mentioning that, for visual simplicity, the schematic shown in Figure \ref{fig:eql_architecture}, shows an EQL network with only two hidden layers, where each layer has only five primitive functions, i.e., the activation function 
\begin{align}
f(\mathbf{g}) \in \big\{\text{identity, square, sine, exponential, multiplication}\big\}\notag.
\end{align}
However, the EQL network can in fact include other functions or more hidden layers to fit a broader range (or class) of functions. Indeed, the number of hidden layers can dictate the complexity of the resulting symbolic expression and plays a similar role to the maximum depth of expression trees in genetic programming techniques. Although the EQL network may not offer the same level of generality as traditional symbolic regression methods, it is adequately capable of representing the majority of functions commonly encountered in scientific and engineering contexts. Crucially, the parametrized nature of the EQL network enables efficient optimization via gradient descent (and backpropagation).

After training the EQL network, the identified equation can be directly derived from the network weights. To avoid reaching overly complex symbolic expressions and to maintain interpretability, it is essential to guide the network towards learning the simplest expression that accurately represents the data. In methods based on genetic programming, this simplification is commonly achieved by restricting the number of terms in the expression. For the EQL network, this is attained by applying sparsity regularization to the network weights, which sets as many of these weights to zero as possible (e.g. $L_1$ regularization \citep{tibshirani1996regression}, $L_{0.5}$ regularization \citep{xu20101}). In this work, we use a smoothed $L_{0.5}$ regularization \citep{wu2014batch, fan2014convergence}, which was also adopted by \citet{kim2020integration}.

\section{Invertible Symbolic Regression}
In this section, we delineate the problem setup for inverse problems, and then describe the proposed ISR approach. 
\label{sec:isr} 
\subsection{Problem Specification}
\label{sec:problem_setup} 
In various engineering and natural systems, the theories developed by experts describe how  measurable (or observable) quantities $\mathbf{y}\in \mathbb{R}^{d_{\mathbf{y}}}$ result from the unknown (or hidden) properties $\mathbf{x}\in \mathbb{R}^{d_{\mathbf{x}}}$, known as the \emph{forward process} ${\bf x} \rightarrow {\bf y}$. The goal of inverse prediction is to predict the unknown variables $\mathbf{x}$ from the observable variables $\mathbf{y}$, through the \emph{inverse process} $\bf y \rightarrow \bf x$. As critical information is lost during the forward process (i.e. $d_{\mathbf{x}} \geq d_{\mathbf{y}}$), the inversion is usually intractable. Given that $f^{-1}(\mathbf{y})$ does not yield a uniquely defined solution, an effective inverse model should instead estimate the \emph{posterior} probability distribution $p(\mathbf{x}\,|\,\mathbf{y})$ of the hidden variables $\mathbf{x}$, conditioned on the observed variable $\mathbf{y}$.

\paragraph{Invertible Symbolic Regression (ISR).} Assume we are given a training dataset $\mathcal{D} = \{\mathbf{x}_i, \mathbf{y}_i\}_{i=1}^N$, 
collected using forward model $\mathbf{y}=s(\mathbf{x})$ and  prior $p(\mathbf{x})$.
To counteract the loss of information during the forward process, we introduce latent random variables $\mathbf{z} \in \mathbb{R}^{d_{\mathbf{z}}}$ drawn from a multivariate standard normal distribution, i.e. $\mathbf{z} \sim p_{_{\bm Z}}(\mathbf{z}) =\mathcal{N}(\boldsymbol{0},\boldsymbol{I}_{d_{\mathbf{z}}})$, where $d_{\mathbf{z}} = d_{\mathbf{x}} - d_{\mathbf{y}}$. These latent variables are designed to capture the information related to $\mathbf{x}$ that is not contained in $\mathbf{y}$ \citep{ArdKruRotKot:19}. 
In ISR, we aim to learn a bijective symbolic function $f: \mathbb{R}^{d_{\mathbf{x}}} \rightarrow \mathbb{R}^{d_{\mathbf{y}}} \times \mathbb{R}^{d_{\mathbf{z}}}$ from the space of functions defined by a set of mathematical functions (e.g. $\sin$, $\cos$, $\exp$, $\log$) and arithmetic operations (e.g. $+$, $-$, $\times$, $\div$),
and such that
\begin{align}
    [\mathbf{y},\mathbf{z}]=f(\mathbf{x}) = \big[f_{\mathbf{y}}(\mathbf{x}),f_{\mathbf{z}}(\mathbf{x})\big], \qquad\qquad \mathbf{x}=f^{-1}(\mathbf{y},\mathbf{z}) \label{eq:ISR_bijective}
\end{align}
where $f_{\mathbf{y}}(\mathbf{x}) \approx s(\mathbf{x})$ is an approximation of the forward process $s(x)$. As discussed later, we will learn $f$ (and hence $f^{-1}$) through an invertible symbolic architecture with bi-directional training. The solution of the inverse
problem (i.e. the posterior $p(\mathbf{x}\,|\,\mathbf{y}^*)$) can then be found by calling $f^{-1}$ for a fixed observation $\mathbf{y}^*$ while randomly (and repeatedly) sampling the latent variable $\mathbf{z}$ from the same standard Gaussian distribution.

  \begin{figure}[t]
    \centering    	\includegraphics[scale=0.45,draft=false]{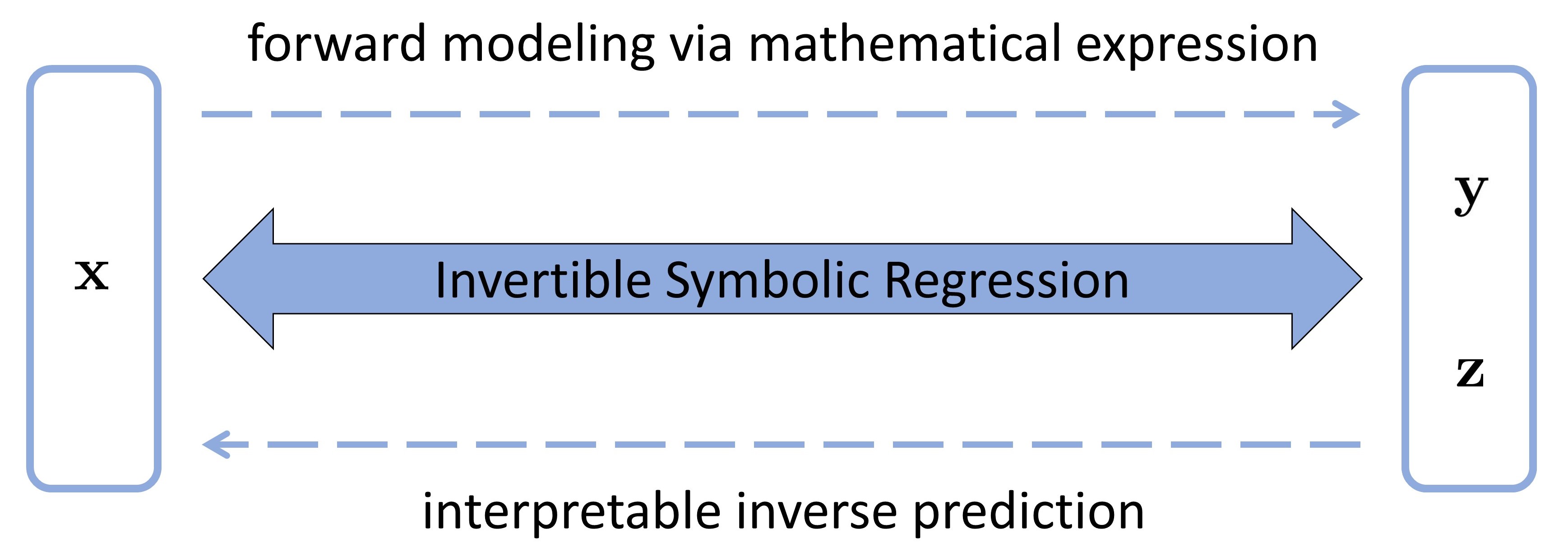}\qquad
    \includegraphics[scale=0.45,draft=false]{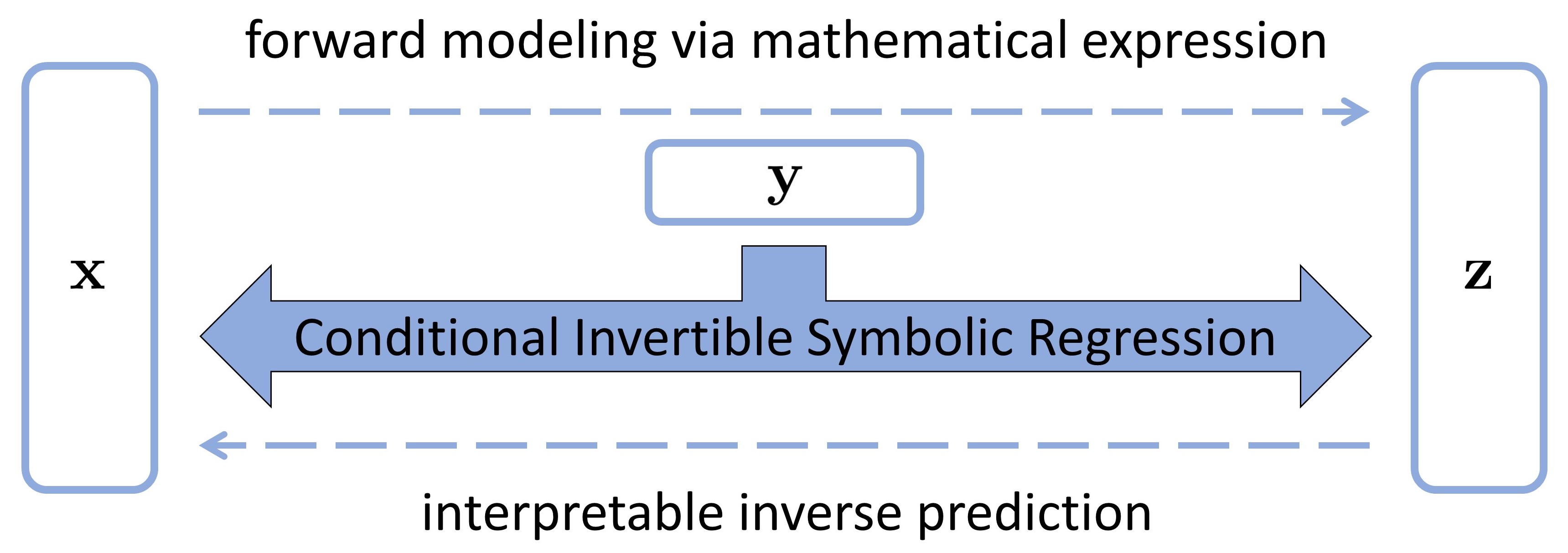}
	\caption{(Left) The proposed ISR framework learns a bijective symbolic transformation that maps the (unknown) variables $\mathbf{x}$ to the (observed) quantities $\mathbf{y}$ while transforming the lost information into latent variables $\mathbf{z}$. (Right) The conditional ISR (cISR) framework learns a bijective symbolic map that transforms $\mathbf{x}$ directly to a latent representation $\mathbf{z}$ given the observation $\mathbf{y}$. As we will show, both the forward and inverse mappings are efficiently computable and possess a tractable Jacobian, allowing explicit computation of posterior probabilities.}
 \label{ISRfigure}
\end{figure} 

\paragraph{Conditional Invertible Symbolic Regression (cISR).} Inspired by works on conditional invertible neural networks (cINNs) \citep{ardizzone2019guided, cINN_ardizone, kruse2021benchmarking,Luce_2023}, instead of training ISR to predict $\mathbf{y}$
from $\mathbf{x}$ while transforming the lost information into latent variables $\mathbf{z}$, we train them to transform $\mathbf{x}$ directly to latent variables $\mathbf{z}$ given the observed variables $\mathbf{y}$. This is achieved by incorporating $\mathbf{y}$ as an additional input within the bijective symbolic architecture during both the forward and inverse passes (see Figure \ref{ISRfigure}). cISR works with larger latent spaces than ISR since $d_{\mathbf{z}} = d_{\mathbf{x}}$ regardless of the dimension $d_{\mathbf{y}}$ of the observed quantities $\mathbf{y}$. Further details are provided in the following section.

In addition to approximating the forward model via mathematical relations, ISR also identifies an interpretable inverse map via analytical expressions (see Figure~\ref{ISRfigure}).
Such interpretable mappings are of particular interest in physical sciences, where an ambitious objective involves creating intelligent machines capable of generating novel scientific findings\citep{udrescu2020aifeynman,udrescu2020aifeynman_2,liu2021machine,keren2023computational,liu2024kan}. As described next, the ISR architecture is both easily invertible and has a tractable Jacobian, allowing for explicit computation of posterior probabilities. 


\subsection{Invertible Symbolic Architecture}

  \begin{figure}[t]
  \noindent 
    \centering    	\includegraphics[scale=0.86,draft=false]{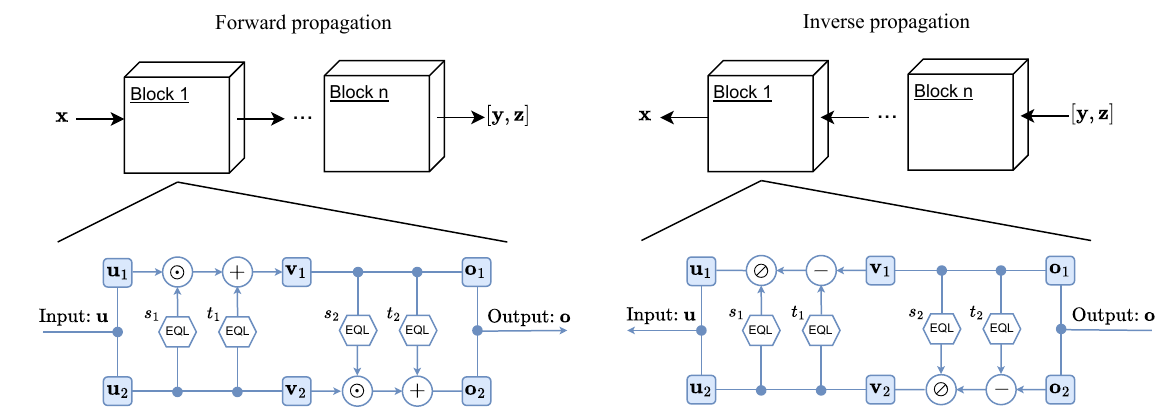}
	\caption{The proposed ISR method integrates EQL within the affine coupling blocks of the INN invertible architecture.\protect\footnotemark \,\,This results in a bijective symbolic transformation that is both easily invertible and has a tractable Jacobian. Indeed, the forward and inverse directions both possess identical computational cost. Here, $\odot$ and $\oslash$ denote element-wise multiplication and divison, respectively.}
\label{fig:coupling_block}
\end{figure}

Inspired by the architectures proposed by~\cite{dinh2016density, kingma2018glow, ArdKruRotKot:19}, we adopt a fully invertible architecture mainly defined by a sequence of $n$ reversible blocks where each block consists of two complementary affine coupling layers. In particular, we first split the block's input $\mathbf{u} \in \mathbb{R}^{d_{\mathbf{u}}}$ into $\mathbf{u}_1 \in \mathbb{R}^{d_{\mathbf{u}_1}}$ and $\mathbf{u}_2 \in \mathbb{R}^{d_{\mathbf{u}_2}}$ (where $d_{\mathbf{u}_1} + d_{\mathbf{u}_2} = d_{\mathbf{u}}$), which are fed into the coupling layers as follows: 
\begin{align}
    \begin{bmatrix}
\mathbf{v}_1\\[\smallskipamount]
\mathbf{v}_2
\end{bmatrix} = 
\begin{bmatrix}
\mathbf{u}_1 \odot \exp\scalebox{1.1}{$($}s_1(\mathbf{u}_2)\scalebox{1.1}{$)$} + t_1(\mathbf{u}_2) \\[\smallskipamount]
\mathbf{u}_2
\end{bmatrix},
\qquad\qquad 
\begin{bmatrix}
\mathbf{o}_1\\[\smallskipamount]
\mathbf{o}_2
\end{bmatrix} = 
\begin{bmatrix}
\mathbf{v}_1 \\[\smallskipamount]
\mathbf{v}_2 \odot \exp\scalebox{1.1}{$($}s_2(\mathbf{v}_1)\scalebox{1.1}{$)$} + t_2(\mathbf{v}_1)
\end{bmatrix},
\label{eq:coupling_layers}
\end{align}
where $\odot$ denotes the Hadamard product or element-wise multiplication. The outputs $[\mathbf{o}_1, \mathbf{o}_2]$ are then concatenated again and passed to the next coupling block. The internal mappings $s_1$ and $t_1$ are functions from $\mathbb{R}^{d_{\mathbf{u}_2}} \rightarrow \mathbb{R}^{d_{\mathbf{u}_1}}$, and $s_2$ and $t_2$ are functions from $\mathbb{R}^{d_{\mathbf{u}_1}} \rightarrow \mathbb{R}^{d_{\mathbf{u}_2}}$. In general, $s_i$ and $t_i$ can be arbitrarily complicated functions (e.g. neural networks as in \cite{ArdKruRotKot:19}). In our proposed ISR approach, they are represented by EQL networks (see Figure \ref{fig:coupling_block}), resulting in a fully symbolic invertible architecture. Moving forward, we shall refer to them as EQL \emph{subnetworks} of the block.
\footnotetext{As direct division can lead to numerical issues, we apply the exponential function to $s_i$ (after clipping its extreme values) in the formulation described in Eq. (\ref{eq:coupling_layers}). This also guarantees non-zero diagonal entries in the Jacobian matrices.}

The transformations above result in upper and lower triangular Jacobians:
\begin{align}
J_{\mathbf{u} \,\mapsto\, \mathbf{v}} =
\begin{bmatrix}
\text{diag}\scalebox{1.2}{$($}\exp\scalebox{1.1}{$($}s_1(\mathbf{u}_2)\scalebox{1.1}{$)$}\scalebox{1.2}{$)$} & \frac{\partial \mathbf{v_1}}{\partial \mathbf{u}_2} \\[\smallskipamount]
0 & I
\end{bmatrix},\qquad\qquad
J_{\mathbf{v} \,\mapsto\, \mathbf{o}} =
\begin{bmatrix}
I &  0 \\[\smallskipamount]
\frac{\partial \mathbf{o_2}}{\partial \mathbf{v}_1} & 
\text{diag}\scalebox{1.2}{$($}\exp\scalebox{1.1}{$($}s_2(\mathbf{v}_1)\scalebox{1.1}{$)$}\scalebox{1.2}{$)$}
\end{bmatrix}.
\end{align}
Hence, their determinants can be trivially computed:
\begin{align}
\mbox{det}\big(J_{\mathbf{u} \,\mapsto\, \mathbf{v}}\big)&= \textstyle\prod_{i=1}^{d_{\mathbf{u}_1}} \exp\scalebox{1.2}{$($}\left[s_1(\mathbf{u}_2)\right]_i\scalebox{1.2}{$)$} = \exp\scalebox{1.7}{$($}\textstyle\sum_{i=1}^{{d_{\mathbf{u}_1}}}\left[s_1(\mathbf{u}_2)\right]_i\scalebox{1.7}{$)$},\notag\\[\smallskipamount]
\mbox{det}\big(J_{\mathbf{v} \,\mapsto\, \mathbf{o}}\big)&= \textstyle\prod_{i=1}^{{d_{\mathbf{u}_2}}} \exp\scalebox{1.2}{$($}\left[s_2(\mathbf{v}_1)\right]_i\scalebox{1.2}{$)$} = \exp\scalebox{1.7}{$($}\textstyle\sum_{i=1}^{{d_{\mathbf{u}_2}}}\left[s_2(\mathbf{v}_1)\right]_i\scalebox{1.7}{$)$}~.
\end{align}
Then, the resulting Jacobian determinant of the coupling block is given by
\begin{align}
\mbox{det}\big(J_{\mathbf{u} \,\mapsto\, \mathbf{o}}\big) &= \mbox{det}\big(J_{\mathbf{u} \,\mapsto\, \mathbf{v}}\big)\cdot\mbox{det}\big(J_{\mathbf{v} \,\mapsto\, \mathbf{o}}\big)\notag\\
&= \exp\scalebox{1.7}{$($}\textstyle\sum_{i=1}^{{d_{\mathbf{u}_1}}}\left[s_1(\mathbf{u}_2)\right]_i\scalebox{1.7}{$)$}\cdot \exp\scalebox{1.7}{$($}\textstyle\sum_{i=1}^{{d_{\mathbf{u}_2}}}\left[s_2(\mathbf{v}_1)\right]_i\scalebox{1.7}{$)$}\notag\\
&= \exp\scalebox{1.7}{$($}\textstyle\sum_{i=1}^{{d_{\mathbf{u}_1}}}\left[s_1(\mathbf{u}_2)\right]_i + \textstyle\sum_{i=1}^{{d_{\mathbf{u}_2}}}\left[s_2(\mathbf{v}_1)\right]_i\scalebox{1.7}{$)$}\notag\\
&= \exp\scalebox{1.7}{$($}\textstyle\sum_{i=1}^{{d_{\mathbf{u}_1}}}\left[s_1(\mathbf{u}_2)\right]_i + \textstyle\sum_{i=1}^{{d_{\mathbf{u}_2}}}\scalebox{1.4}{$[$}s_2\scalebox{1.3}{$($}\mathbf{u}_1 \odot \exp\scalebox{1.1}{$($}s_1(\mathbf{u}_2)\scalebox{1.1}{$)$} + t_1(\mathbf{u}_2)\scalebox{1.3}{$)$}\scalebox{1.4}{$]$}_i\scalebox{1.7}{$)$}
\end{align}
which can be efficiently calculated. Indeed, the Jacobian determinant of the whole map $\mathbf{x} \rightarrow [\mathbf{y}, \mathbf{z}]$ is the product of the Jacobian determinants of the $n$ underlying coupling blocks (see Figure \ref{fig:coupling_block}). 

Given the output $\mathbf{o} = [\mathbf{o}_1, \mathbf{o}_2]$, the expressions in Eqs. (\ref{eq:coupling_layers}) are clearly invertible:
\begin{align}
\mathbf{u}_2 = \big(\mathbf{o}_2 - t_2(\mathbf{o}_1)\big) \oslash \exp \scalebox{1.1}{$($} s_2(\mathbf{o}_1) \scalebox{1.1}{$)$}, \qquad\qquad
\mathbf{u}_1 = \big(\mathbf{o}_1 - t_1(\mathbf{u}_2)\big) \oslash  \exp \scalebox{1.1}{$($} s_1(\mathbf{u}_2) \scalebox{1.1}{$)$}
\label{eq:inverse_coupling_layers}
\end{align}
where $\oslash$ denotes element-wise division. Crucially, even when the coupling block is inverted, the EQL subnetworks $s_i$ and $t_i$ need \emph{not} themselves be invertible; they are only ever evaluated in the forward direction. We denote the whole ISR map $\mathbf{x} \rightarrow [\mathbf{y},\mathbf{z}]$ as $f(\mathbf{x};\theta) = \big[f_{\mathbf{y}}(\mathbf{x};\theta),f_{\mathbf{z}}(\mathbf{x};\theta)\big]$ parameterized by the EQL subnetworks parameters $\theta$, and the inverse as $f^{-1}(\mathbf{y}, \mathbf{z}; \theta)$.\vspace{5pt}

\remark The proposed ISR architecture consists of a sequence of these symbolic reversible blocks. To enhance the model's predictive and expressive capability, we can: i) increase the number of reversible coupling blocks, ii) increase the number of hidden layers in each underlying EQL network, iii) increase the number of hidden neurons per layer in each underlying EQL network, or iv) increase the complexity of the symbolic activation functions used in the EQL network. However, it is worth noting that these enhancements come with a trade-off, as they inevitably lead to a decrease in the model's interpretability.\vspace{5pt}

\remark To further improve the model capacity, as in \citet{ArdKruRotKot:19}, we incorporate (random, but fixed) permutation layers between the coupling blocks, which shuffles the input elements for subsequent coupling blocks. This effectively randomizes the configuration of splits $\mathbf{u} = [\mathbf{u}_1, \mathbf{u}_2]$ across different blocks, thereby enhancing interplay between variables.\vspace{5pt}

\remark Inspired by \citet{ArdKruRotKot:19}, we split the coupling block's input vector $\mathbf{u} \in \mathbb{R}^{d_{\mathbf{u}}}$ into two halves, i.e. $\mathbf{u}_1 \in \mathbb{R}^{d_{\mathbf{u}_1}}$ and $\mathbf{u}_2 \in \mathbb{R}^{d_{\mathbf{u}_2}}$ where $d_{\mathbf{u}_1} = \floor{\frac{d_{\mathbf{u}}}{2}}$ and $d_{\mathbf{u}_2} = d_{\mathbf{u}} - d_{\mathbf{u}_1}$. In the case where $\mathbf{u}$ is one-dimensional (or scalar), i.e. $d_{\mathbf{u}} = 1$ and $\mathbf{u} \in \mathbb{R}$, we pad it with an extra zero (so that $d_{\mathbf{u}} = 2$) along with a loss term that prevents the encoding of information in the extra dimension (e.g. we use the  $L_2$ loss to maintain those values near zero).\vspace{5pt} 

\remark The proposed ISR architecture is also compatible with the conditional ISR (cISR) framework proposed in the previous section. In essence, cISR identifies a bijective symbolic transformation directly between $\mathbf{x}$ and $\mathbf{z}$ given the observation $\mathbf{y}$. This is attained by feeding $\mathbf{y}$ as an extra input to each coupling block, during both the forward and inverse passes. In particular, and as suggested by \citet{ardizzone2019guided, cINN_ardizone, kruse2021benchmarking}, we adapt the same coupling layers given by Eqs.~(\ref{eq:coupling_layers}) and (\ref{eq:inverse_coupling_layers}) to produce a conditional coupling block. Since the subnetworks $s_i$ and $t_i$ are never inverted, we enforce the condition on the observation by concatenating $\mathbf{y}$ to their inputs without losing the invertibility, i.e. we replace $s_1(\mathbf{u}_2)$
with $s_1(\mathbf{u}_2, \mathbf{y})$, etc. In complex settings, the condition $\mathbf{y}$ is first fed into a separate feed-forward conditioning network, resulting in higher-level conditioning features that are then injected into the conditional coupling blocks. Although cISR can have better generative properties \citep{kruse2021benchmarking}, it leads to more complex symbolic expressions and less interpretability as it explicitly conditions the map on the observation within the symbolic formulation. We denote the entire cISR forward map $\mathbf{x} \rightarrow \mathbf{z}$ as $f(\mathbf{x}; \mathbf{y}, \theta)$ parameterized by $\theta$, and the inverse as $f^{-1}(\mathbf{z}; \mathbf{y}, \theta)$.

\subsection{Maximum Likelihood Training of ISR} 
\label{sec:MLE_loss}
We train the proposed ISR model to learn a bijective symbolic transformation $f: \mathbb{R}^{d_{\mathbf{x}}} \rightarrow \mathbb{R}^{d_{\mathbf{y}}} \times \mathbb{R}^{d_{\mathbf{z}}}$. There are various choices to define the loss functions with different advantage and disadvantages~\citep{grover2018flow, ren2020benchmarking, ArdKruRotKot:19,cINN_ardizone,kruse2021benchmarking}. As reported in \cite{kruse2021benchmarking}, there are two main training approaches: \vspace{5pt}\\
i) A standard supervised $L_2$ loss for fitting the model's $\mathbf{y}$ predictions to the training data, combined with a Maximum Mean Discrepancy (MMD) \citep{gretton2012kernel,ArdKruRotKot:19} for fitting the latent distribution $p_{_{\bm Z}}(\mathbf{z})$ to $\mathcal{N}(\boldsymbol{0},\boldsymbol{I}_{d_{\mathbf{z}}})$, given samples.\vspace{5pt}\\
ii) A Maximum Likelihood Estimate (MLE) loss that enforces $\mathbf{z}$ to be standard Gaussian, i.e. $\mathbf{z} \sim p_{_{\bm Z}}(\mathbf{z}) =\mathcal{N}(\boldsymbol{0},\boldsymbol{I}_{d_{\mathbf{z}}})$ and by approximating the distribution on $\mathbf{y}$ with a Gaussian distribution around the ground truth values $\mathbf{y}_{\text{gt}}$ with very low variance $\sigma^2$
\citep{dinh2016density, ren2020benchmarking, kruse2021benchmarking}. 

Given that MLE is shown to perform well as reported in the literature \citep{ArdKruRotKot:19}, we apply it here. Next, we demonstrate how this approach is equivalent to minimizing the forward Kullback-Leibler (KL) divergence as the cost (cf.~\citep{papamakarios2021normalizing}).
We note that given the map $f(\mathbf{x};\theta)\mapsto [\mathbf{z},\mathbf{y}]$, parameterized by $\theta$, and assuming $\mathbf{y}$ and $\mathbf{z}$ are independent, the density $p_{_{\mathbf{X}}}$ relates to $p_{_{\mathbf{Y}}}$ and  $p_{_{\mathbf{Z}}}$ through the change-of-variables formula
\begin{align}
\label{eq:change_var_4_KL}
p_{_{\bm X}}(\mathbf{x}; \theta) = p_{_{\bm Y}}\big(\mathbf{y}=f_\mathbf{y}(\mathbf x; \theta)\big)\,\, p_{_{\bm Z}}\big(\mathbf{z} = f_{\mathbf{z}}(\mathbf{x};\theta)\big)\cdot \left|\mbox{det}\big(J_{\mathbf{x} \,\mapsto\,[\mathbf{z},\mathbf{y}]}(\mathbf{x};\theta)\big)\right|.
\end{align}
where $J_{\mathbf{x} \,\mapsto\,[\mathbf{z},\mathbf{y}]}(\mathbf{x};\theta)$ denotes the Jacobian of the map $f$ parameterized by $\theta$. This expression is then used to define the loss function, which we derive by following the work in \citep{papamakarios2021normalizing}. In particular, we aim to minimize the forward KL divergence between a target distribution $p^*_{_{\bm X}}(\mathbf{x})$ and our \emph{flow-based} model $p_{_{\bm X}}(\mathbf{x}; \theta)$, given by
\begin{align}
\mathcal{L}(\theta) &= D_{\text{KL}}\big[p^*_{_{\bm X}}(\mathbf{x})\,\big|\big|\,p_{_{\bm X}}(\mathbf{x}; \theta)\big]\notag\\
&= - \mathbb{E}_{p^*_{\bm X}(\mathbf{x})}\big[\log p_{_{\bm X}}(\mathbf{x}; \theta)\big] + \text{const.}\notag\\
&= - \mathbb{E}_{p^*_{\bm X}(\mathbf{x})}\big[
\log p_{_{\bm Y}}\big(f_\mathbf{y}(\mathbf x; \theta)\big) + 
\log p_{_{\bm Z}}\big(f_{\mathbf{z}}(\mathbf{x};\theta)\big) +
\log\left|\mbox{det}\big(J_{\mathbf{x} \,\mapsto\,[\mathbf{z},\mathbf{y}]}(\mathbf{x};\theta)\big)\right|
\big] + \text{const.}\label{eq:kl_div}
\end{align}
The forward KL divergence is particularly suitable for cases where we have access to samples from the target distribution, but we cannot necessarily evaluate the target density $p^*_{_{\bm X}}(\mathbf{x})$. Assuming we have a set of samples $\{\mathbf{x}_i\}_{i=1}^N$ from $p^*_{_{\bm X}}(\mathbf{x})$, we can approximate the expectation in Eq.~(\ref{eq:kl_div}) using Monte Carlo integration as
\begin{align}
\label{eq:KL_forward_app}
\mathcal{L}(\theta) &\approx - \frac{1}{N}\sum_{i=1}^N \Big( 
\log p_{_{\bm Y}}\big(f_\mathbf{y}(\mathbf x_i; \theta)\big) + 
\log p_{_{\bm Z}}\big(f_{\mathbf{z}}(\mathbf{x}_i;\theta)\big) +
\log\left|\mbox{det}\big(J_{\mathbf{x} \,\mapsto\,[\mathbf{z},\mathbf{y}]}(\mathbf{x}_i;\theta)\big)\right|
+ \text{const.}\Big) ~.
\end{align}
As we can see, minimizing the above Monte Carlo approximation of the KL divergence is equivalent to maximizing likelihood (or minimizing negative log-likelihood). Assuming $p_{_{\mathbf{Z}}}$ is standard Gaussian and $p_{_{\mathbf{Y}}}$ is a multivariate normal distribution around $\mathbf{y}_\mathrm{gt}$, the negative log-likelihood (NLL) loss in Eq.~(\ref{eq:KL_forward_app}) becomes
\begin{align}
\mathcal{L}_\mathrm{NLL}(\theta) &= \frac{1}{N}\sum_{i=1}^N \left( 
\frac{1}{2}\cdot\frac{\big(f_\mathbf{y}(\mathbf{x}_i; \theta) - \mathbf{y}_{\text{gt}}\big)^2}{\sigma^2} + \frac{1}{2}\cdot f_{\mathbf{z}}(\mathbf{x}_i;\theta)^2 - \log\left|\mbox{det}\big(J_{\mathbf{x} \,\mapsto\,[\mathbf{z},\mathbf{y}]}(\mathbf{x}_i;\theta)\big)\right|\right) \label{eq:nll_new}~.
\end{align}
In other words, we find the optimal ISR parameters $\theta$ by minimizing the NLL loss in Eq.~(\ref{eq:nll_new}), and the resulting bijective symbolic expression can be directly extracted from the these optimal parameters.\vspace{5pt}

\remark We note that cISR is also suited for maximum likelihood training. Given the conditioning observation $\mathbf{y}$, the density $p_{_{\mathbf{X}\,|\,\mathbf{Y}}}$ relates to $p_{_{\mathbf{Z}}}$ through the change-of-variables formula 
\begin{align}
p_{_{\mathbf{X}\,|\,\mathbf{Y}}}(\mathbf{x}\,|\,\mathbf{y}, \theta) = p_{_{\mathbf{Z}}}\big(\mathbf{z} = f(\mathbf{x}; \mathbf{y},\theta)\big) \cdot \left|\mbox{det}\big(J_{\mathbf{x} \,\mapsto\,\mathbf{z}}(\mathbf{x};\mathbf{y},\theta)\big)\right|,
\end{align}
where $J_{\mathbf{x} \,\mapsto\,\mathbf{z}}(\mathbf{x};\mathbf{y},\theta)$ indicates the Jacobian of the map $f$ conditioned on $\mathbf{y}$ and parameterized by $\theta$. Following the same procedure as above, the cISR model can be trained by minimizing the following NLL loss function
\begin{align}
\label{eq:NLL_cost_4_cISR}
\mathcal{L}_\mathrm{NLL}(\theta) &= \frac{1}{N}\sum_{i=1}^N \left( 
\frac{1}{2}\cdot f(\mathbf{x}_i;\mathbf{y}_i, \theta)^2 - \log\left|\mbox{det}\big(J_{\mathbf{x} \,\mapsto\,\mathbf{z}}(\mathbf{x}_i;\mathbf{y}_i, \theta)\big)\right|\right)~.
\end{align}
As we will show in the next section, if we ignore the condition on the observation $\mathbf{y}$, the loss in Eq.~\ref{eq:NLL_cost_4_cISR} can also be used for training ISR as a normalizing flow for the unsupervised learning task of approximating a target probability density function from samples (cf. Eq.~\ref{eq:NNLcostforNF}).  

\section{Results}
\label{sec:results}
We evaluate our proposed ISR method on a variety of problems. We first show how ISR can serve as a normalizing flow for density estimation tasks on several test distributions. We then demonstrate the capabilities of ISR in solving inverse problems by considering two synthetic problems and then a more challenging application in ocean acoustics~\citep{jensen2011computational,ali2023mseas,huang2006uncertainty,DosDet:11,BiaGerTra:19,holland2005remote, benson2000geoacoustic}. We mainly compare our ISR approach against INN \citep{ArdKruRotKot:19} throughout our experiments. Further experimental details can be found in Appendix \ref{app:b}.

\subsection{Leveraging ISR for Density Estimation via Normalizing Flow}
\label{sec:density_estimation_normalizing_flow}

Given $N$ independently and identically distributed (i.i.d.) samples, i.e. $\{\textbf{X}_i\}_{i=1}^N\sim p^\mathrm{target}_{_{\mathbf{X}}}$, we would like to estimate the target density $p^\mathrm{target}_{_{\mathbf{X}}}$ and generate new samples from it. This problem categorizes as the density estimation where non-parametric, e.g. Kernel Density Estimation \citep{sheather2004density}, and parametric estimators, e.g. Maximum Entropy Distribution as the least biased estimator \citep{tohme2024messy}, are classically used. In recent years, this problem has been approached using normalizing flow equipped with invertible map, which has gained a great deal of interest in the generative AI task. In an attempt to introduce interpretability in the trained model, we extend invertible normalizing flow to symbolic framework using the proposed ISR architecture.

In order to use the proposed ISR method as the normalizing flow for the unsupervised task of resampling from an intractable target distribution, we drop out $\mathbf{y}$ and enforce $d_{\mathbf{x}} = d_{\mathbf{z}}$.\footnote{In the absence of $\mathbf{y}$, cISR and ISR are equivalent, so we simply refer to them as ISR. Similarly, INN and cINN become equivalent, and we simply refer to them as INN.} In this case, we aim to learn a an invertible and symbolic map $f: \mathbb{R}^{d_{\mathbf{x}}} \rightarrow \mathbb{R}^{d_{\mathbf{z}}}$, parameterized by $\theta$ such that 
\begin{align}
    \mathbf{z}=f(\mathbf{x}; \theta), \qquad\qquad \mathbf{x}=f^{-1}(\mathbf{z}; \theta),\label{eq:ISR_bijective}
\end{align}
where $\textbf z \sim p_{_{\mathbf{Z}}}$ is the standard normal distribution function, which is easy to sample from. 
Using the change-of-variables formula, the density $p_{_{\mathbf{X}}}$ relates to the density $p_{_{\mathbf{Z}}}$ via
\begin{align}
p_{_{\mathbf{X}}}(\mathbf{x};\theta) = p_{_{\mathbf{Z}}}\big(\mathbf{z}=f(\mathbf{x}; \theta)\big) \cdot \left|\mbox{det}\big(J_{\mathbf{x} \,\mapsto\,\mathbf{z}}(\mathbf{x};\theta)\big)\right|,
\end{align}
where $J_{\mathbf{x} \,\mapsto\,\mathbf{z}}$ indicates the Jacobian of the map $f$ parameterized by $\theta$. Following the same procedure outlined in Section \ref{sec:MLE_loss}, the model can be trained by minimizing the following NLL loss function
\begin{align}
\label{eq:NNLcostforNF}
\mathcal{L}_\mathrm{NLL}(\theta) &= \frac{1}{N}\sum_{i=1}^N \left( 
\frac{1}{2}\cdot f(\mathbf{x}_i; \theta)^2 - \log\left|\mbox{det}\big(J_{\mathbf{x} \,\mapsto\,\mathbf{z}}(\mathbf{x}_i; \theta)\big)\right|\right)~.
\end{align}
\begin{figure}[t]
    \centering    	
    \includegraphics[scale=0.6]{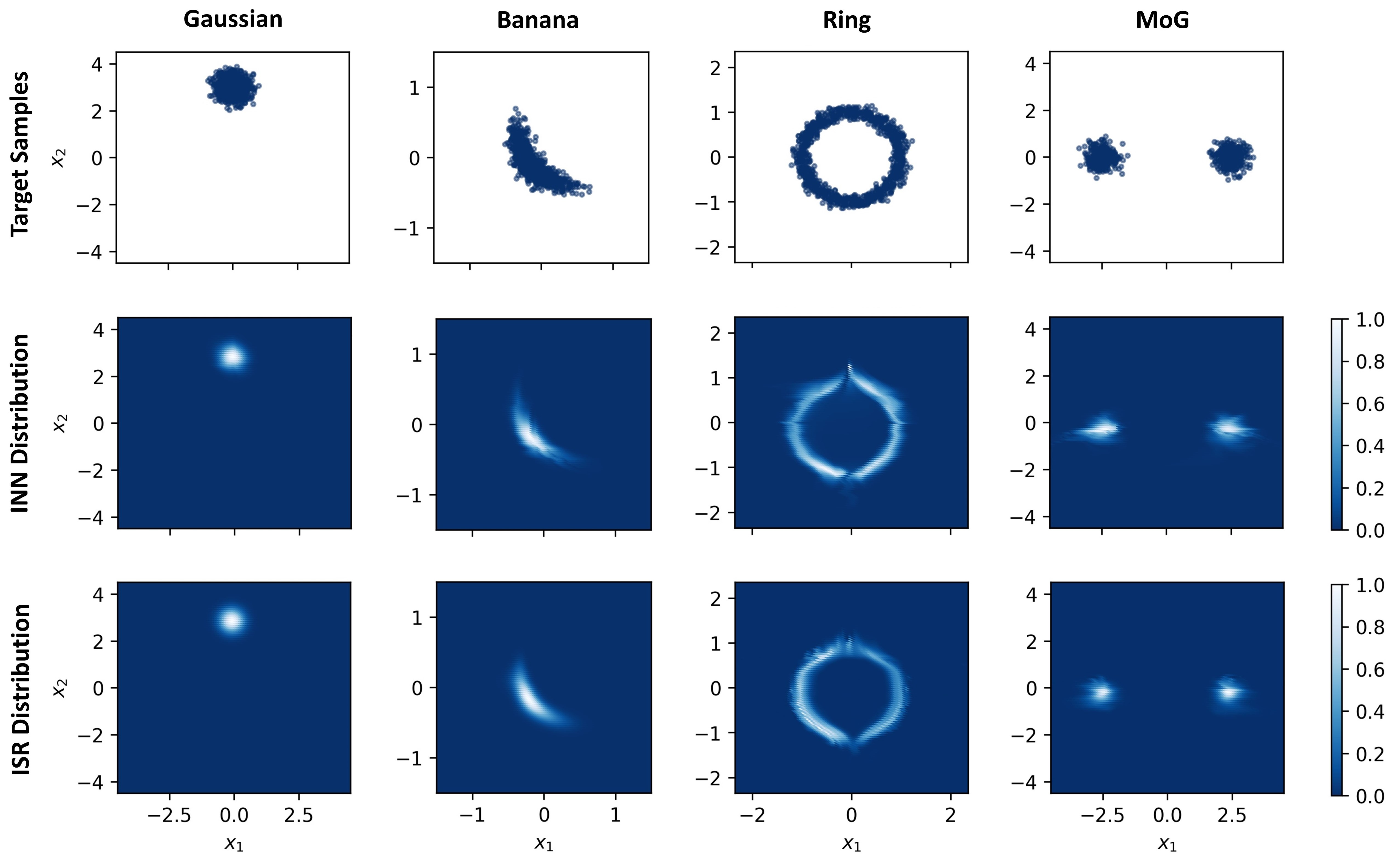}
	\caption{Samples from four different target densities (first row), and their estimated distributions using INN (second row) and the proposed ISR method (third row).}
 \label{fig:normalizing_flow}
 \vspace{-0.29cm}
\end{figure}

We compare the proposed ISR approach with INN in recovering several two-dimensional target distributions (i.e. $d_{\mathbf{x}} = d_{\mathbf{z}} = 2)$. First, we consider a fairly simple multivariate normal distribution $\mathcal{N}\big(\boldsymbol{\mu}, \boldsymbol{\Sigma}\big)$ with mean $\boldsymbol{\mu} = [0,3]$ and covariance matrix $\boldsymbol{\Sigma} = \frac{1}{10}\cdot \boldsymbol{I}_2$ as the target density. Then, we consider more challenging distributions: the ``Banana,'' ``Mixture of Gaussians (MoG),'' and ``Ring'' distributions that are also considered in \citet{jaini2019sum, wenliang2019learning}. For each of these target distributions, we draw $N_s=10^4$ i.i.d. samples and train an invertible map that transport the samples to a standard normal distribution. This is called normalizing flow, where we intend to compare the standard INN with the proposed ISR architecture. Here, we use a single coupling block for the Gaussian and banana cases, and two coupling blocks for the ring and MoG test cases. 
\\ \ \\
As shown in Figure \ref{fig:normalizing_flow}, the proposed ISR method finds and generates samples of the considered target densities with slightly better accuracy than INN. We report the bijective symbolic expressions in Table \ref{table:isr_expr} of Appendix \ref{app:a}. For instance, the first target distribution in Figure \ref{fig:normalizing_flow} is the two-dimensional multivariate Gaussian distribution $\mathcal{N}\big(\boldsymbol{\mu}, \boldsymbol{\Sigma}\big)$ with mean $\boldsymbol{\mu} = [0,3]$ and covariance matrix $\boldsymbol{\Sigma} = \frac{1}{10}\cdot \boldsymbol{I}_2$. This is indeed a shifted and scaled standard Gaussian distribution where we know the analytical solution to the true map:
\begin{align}
    \mathbf{X} = \begin{bmatrix}
X_1\\
X_2
\end{bmatrix} \sim \mathcal{N}\left(\bm \mu, \frac{1}{10}\cdot\bm I_2\right) 
    &\sim \bm \mu + \sqrt{\frac{1}{10}} \cdot \mathcal{N}\left(\bm 0, \bm I_2\right)
    \notag\\
    &= \bm \mu + \frac{1}{\sqrt{10}} \cdot \mathbf{Z}
    =\begin{bmatrix}
0\\
3
\end{bmatrix} + 0.316 \cdot \begin{bmatrix}
Z_1\\
Z_2
\end{bmatrix}
=
 \begin{bmatrix}
0.316\,Z_1\\
3 + 0.316\,Z_2
\end{bmatrix}~.
\label{eq:gaussian_true_map}
\end{align}

As shown in Table \ref{table:isr_expr} of Appendix \ref{app:a}, for this Gaussian distribution example, the proposed ISR method finds the following bijective expression:
\begin{align}
z_1 &= x_1 \cdot e^{1.16} = 3.19\,x_1\hspace{2.81cm} \Longleftrightarrow \qquad x_1 = 3.19^{-1}\, z_1 = 0.313 \,z_1\notag\\
z_2 &= x_2 \cdot e^{1.14} - 9.39 = 3.13\,x_2 - 9.39\qquad \Longleftrightarrow \qquad x_2 = 3.13^{-1}\, (z_2 + 9.39) = 0.319\, z_2 + 3.00\notag
\end{align}
In other words, the proposed ISR method identifies the true underlying transformation given by Eq. (\ref{eq:gaussian_true_map}) with a high accuracy. 



\subsection{Inverse Kinematics}
\label{sec:inverse_kinematics}
\begin{figure}[t]
\centering    	
\includegraphics[scale=0.65]{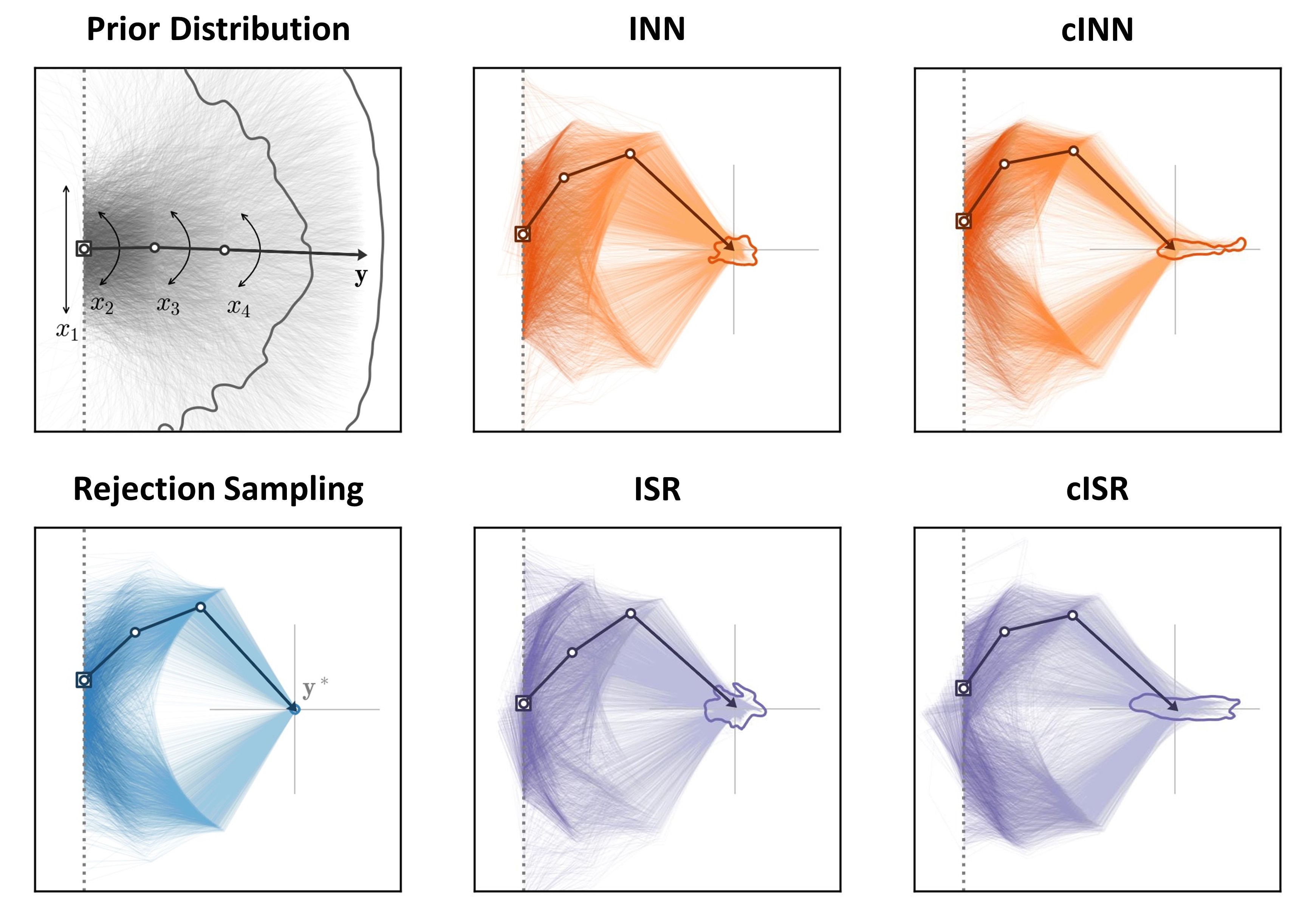}
\caption{Results for the inverse kinematics benchmark problem. The faint colored lines indicate sampled arm configurations $\mathbf{x}$ taken from each model's predicted posterior $\hat{p}(\mathbf{x}\,|\,\mathbf{y}^*)$, conditioned on the target end point $\mathbf{y}^*$, which is indicated by a gray cross. The contour lines around the target end point enclose the regions containing 97\% of the sampled arms’ end points. We emphasize the arm with the highest estimated likelihood as a bold line.
}
 \label{fig:inverse_kinematics}
\end{figure}
We now consider a geometrical benchmark example used by \citet{ArdKruRotKot:19, kruse2021benchmarking}, which simulates an inverse kinematics problem in a two-dimensional space: A multi-jointed 2D arm moves vertically along a rail and rotates at three joints. In this problem, we are interested in the configurations (i.e. the four degrees of freedom) of the arm that place the arm's end point at a given position. The forward process computes the coordinates of the end point $\mathbf{y} \in\mathbf{R}^2$, given a configuration $\mathbf{x}\in \mathbf{R}^4$ (i.e. $d_{\mathbf{x}} = 4$, $d_{\mathbf{y}} = 2$, and hence $d_{\mathbf{z}} = 2$). In particular, the forward process takes $\mathbf{x} = [x_1, x_2, x_3, x_4]$ as argument, where $x_1$ denotes the arm's starting height, and $x_2$, $x_3$, $x_4$ are its three joint angles, and returns the coordinates of its end point $\mathbf{y} = [y_1, y_2]$ given by
\begin{align}
y_1 &= \ell_1\sin(x_2) + \ell_2\sin(x_2 + x_3) + \ell_3 \sin(x_2 + x_3 + x_4) + x_1\notag\\[\smallskipamount]
y_2 &= \ell_1\cos(x_2) + \ell_2\cos(x_2 + x_3) + \ell_3 \cos(x_2 + x_3 + x_4)\label{eq:inverse_kinematics_forward}
\end{align} 
where the segment lengths $\ell_1 = 0.5$, $\ell_2 = 0.5$, and $\ell_3 = 1$. The parameters $\mathbf{x}$ follow a Gaussian prior $\mathbf{x} \sim \mathcal{N}\big(\boldsymbol{0}, \boldsymbol{\sigma}^2\cdot\boldsymbol{I}_{4}\big)$ with $\boldsymbol{\sigma}^2 = [0.25^2, 0.25, 0.25, 0.25]$, which favors a configuration with a centered origin and $180^{\circ}$ joint angles (see Figure \ref{fig:inverse_kinematics}). We consider a training dataset of size $10^6$, constructed using this Gaussian prior and the forward process in Eq.~(\ref{eq:inverse_kinematics_forward}).
The inverse problem here asks to find the posterior distribution $p(\mathbf{x}\,|\,\mathbf{y}^*)$ of all possible configurations (or parameters) $\mathbf{x}$ that result in the arm's end point being positioned at a given $\mathbf{y}^*$ location. This inverse kinematics problem, being low-dimensional, offers computationally inexpensive forward (and backward) process, which enables fast training, intuitive visualizations, and an approximation of the true posterior estimates via rejection sampling.\footnote{\textbf{Rejection sampling.} Assume we require $N_s$ samples of $\mathbf{x}$ from the posterior $p(\mathbf{x}\,|\,\mathbf{y}^*)$ given some observation $\mathbf{y}^*$. After setting some acceptance threshold $\epsilon$, we iteratively generate $\mathbf{x}$-samples from the prior. For each sample, we simulate the corresponding $\mathbf{y}$-values and only keep those with $\text{dist}(\mathbf{y}, \mathbf{y}^*) < \epsilon$. The process is repeated until $N_s$ samples are collected (or accepted). Indeed, the smaller the threshold $\epsilon$, the more $\mathbf{x}$-samples candidates (and hence the more simulations) have to be generated. Hence, we adopt this approach in this low-dimensional inverse kinematics problem, where we can afford to run the forward process (or simulation) a huge number of times.}

An example of a challenging end point $\mathbf{y}^*$ is shown in Figure \ref{fig:inverse_kinematics}, where we compare the proposed ISR method against the approximate true posterior (obtained via rejection sampling), as well as INN. The chosen $\mathbf{y}^*$ is particularly challenging, since this end point is unlikely under the prior $p(\mathbf{x})$, and results in a strongly bi-modal posterior $p(\mathbf{x}\,|\,\mathbf{y}^*)$ \citep{ArdKruRotKot:19, kruse2021benchmarking}. As we can observe in Figure \ref{fig:inverse_kinematics}, compared to rejection sampling, all the considered architectures (i.e. INN, cINN, ISR, and cISR) are able to capture the two symmetric modes well. However, we can clearly see that they all generate $\mathbf{x}$-samples such that their resulting end points miss the target $\mathbf{y}^*$ by a wider margin. Quantitative results are also provided in Appendix \ref{app:c}.

\subsection{Application: Geoacoustic Inversion}
\label{sec:geoacoustic_inversion}
Predicting acoustic propagation at sea is vital for various applications, including sonar performance forecasting and mitigating noise pollution at sea. The ability to predict sound propagation in a shallow water environment depends on understanding the seabed's geoacoustic characteristics. Inferring those characteristics from ocean acoustic measurements (or signals) is known as geoacoustic inversion (GI).
GI involves several components: (i) representation of the ocean environment, (ii) selection of the inversion method, including the forward propagation model implemented, and (iii) quantification of the uncertainty related to the parameters estimates.


 \begin{figure}[t]
    \centering    	\includegraphics[scale=0.65,draft=false]{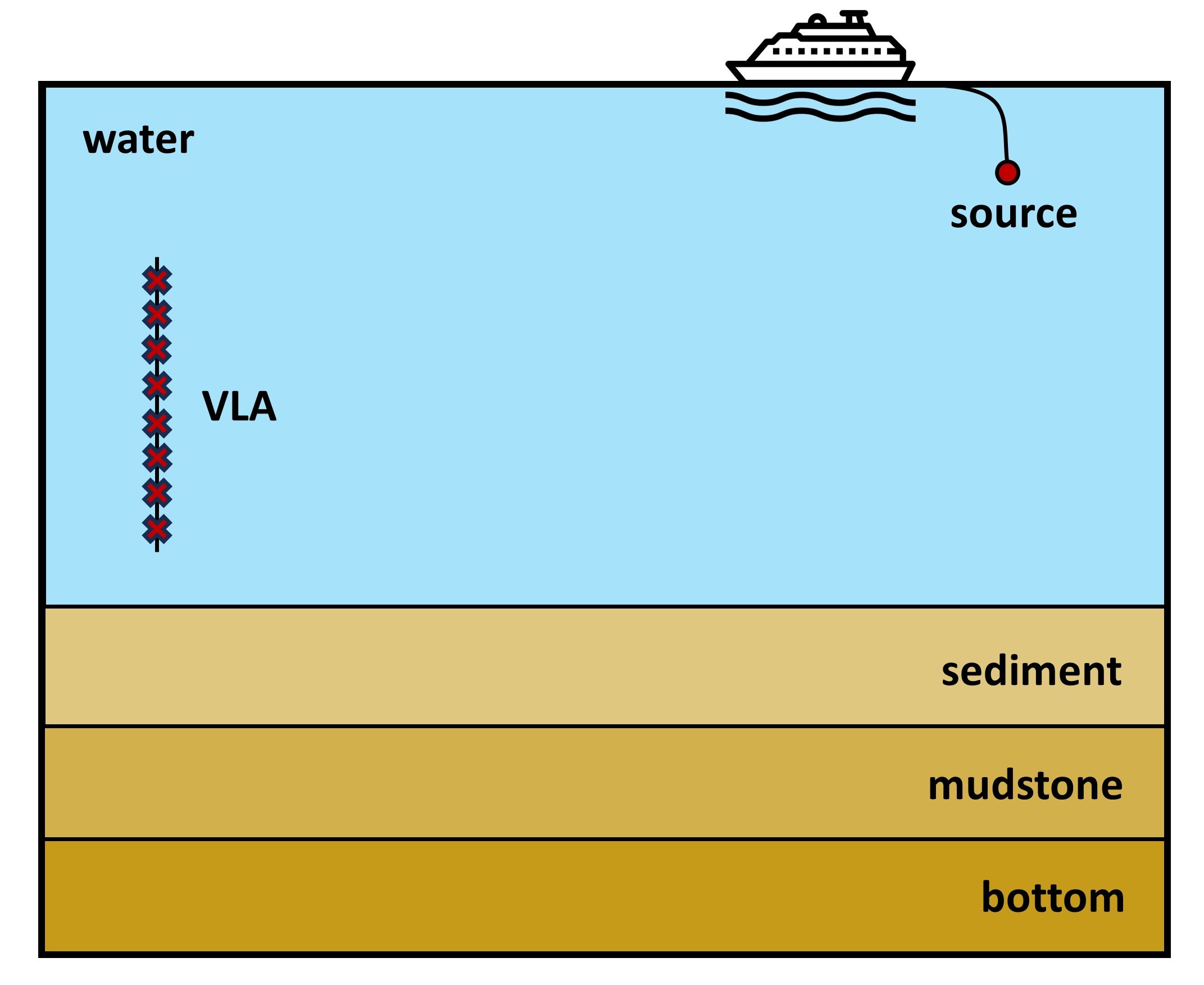}
	\caption{The SWellEx-96 experiment environment. The acoustic source is towed by a research vessel and transmits signals at various frequencies. The acoustic sensor consists of a vertical line array (VLA). Based on the measurements collected at the VLA, the objective is to estimate posterior distributions over parameters of interest (e.g. water depth, sound speed at the water-sediment interface, source range and depth, etc.).}
 \label{SWellEx-96-diagram}
\end{figure}


We start by describing the ocean environment. We consider the setup of 
SWellEx-96~\cite{yardim2010geoacoustic,MeyGem:J21}, which was an experiment done off the coast of San Diego, CA, near Point Loma. This experimental setting  is one of the most used, documented, and understood studies in the undersea acoustics community.\footnote{see \url{http://swellex96.ucsd.edu/}} As depicted in Figure~\ref{SWellEx-96-diagram}, the data is collected via a vertical line array (VLA). 
The specification of the 21 hydrophones of the VLA and sound speed profile (SSP) in the water column is provided in the SWellEx-96 documentation. The SSP and sediment parameters are considered to be range-independent. Water depth refers to the depth of the water at the array. The source is towed by a research vessel which  consists of a comb signal comprising  frequencies of $49,79,112,148,201,283,\ {\mathrm{and}}\ 388\ \mathrm{Hz}$. While in the SWellEx-96 experiment the position of the source changes with time, for this task we consider the instant when the source depth is $60$ m and the distance (or range) between the source and the VLA is \mbox{$3$ km}.
 
 
The sediment layer is modeled with the following properties.
The seabed consists initially of a sediment layer that is $23.5$ meters thick, with a density of $1.76$ g/$\mbox{cm}^3$, and an attenuation of $0.2$ dB/kmHz. The sound speed at the bottom of this layer is assumed to be 
$1593$ m/s. The second layer is mudstone that is $800$ meters thick, possessing a density of $2.06$ g/$\mbox{cm}^3$, and an attenuation of $0.06$ dB/kmHz. The top and bottom sound speeds of this layer are $1881$ m/s and $3245$ m/s respectively. The description of the geoacoustic model of the SWellEx-96 experiment is complemented by a half-space featuring a density of $2.66$ g/$\mbox{cm}^3$, an attenuation of $0.020$ dB/kmHz, and a sound speed of $5200$ m/s. 

Here, we consider two geoacoustic inversion tasks:

\textit{Task 1.} 
Based on the measurements at the VLA, the objective of this task is to infer the posterior distribution over the water depth as well as the sound speed at the water-sediment interface.
For this task, we assume all the quantities above to be known. The unknown parameters $m_1$ (the water depth) and $m_2$ (the sound speed at the water-sediment interface) follow a uniform prior in $[200.5, 236.5]$ m and $[1532, 1592]$ m/s, i.e. \mbox{$m_1\sim \mathcal{U}([200.5, 236.5])$} and $m_2\sim \mathcal{U}([1532, 1592])$, where $\mathcal{U}(\Omega)$ denotes a uniform distribution in \mbox{the domain $\Omega$.}

\textit{Task 2.} In addition to the two parameters considered in \textit{Task 1} (i.e. the water depth $m_1$ and the sound speed at the water-sediment interface $m_2$), we also estimate the posterior distribution over the VLA tilt $m_3$, as well as the thickness of the first (sediment) layer $m_4$. All other quantities provided above are assumed to be known. As in \textit{Task 1}, the unknown parameters follow a uniform prior, i.e. $m_1\sim \mathcal{U}([200.5, 236.5])$, $m_2\sim \mathcal{U}([1532, 1592])$, $m_3 \sim \mathcal{U}([-2, 2])$, and $m_4\sim\mathcal{U}([18.5, 28.5])$.

The received pressure {\bf y} on each hydrophone and for each frequency is a function of unknown parameters ${\bf m}$ (e.g. water depth, sound speed at the water-sediment interface, etc.) and additive noise $\boldsymbol{\epsilon}$ as follows
\begin{align}
\label{eq:GI-forwardmodel}
    {\bf y}= s({\bf m}, \boldsymbol{\epsilon})=F({\bf m}) + \boldsymbol{\epsilon},\qquad\qquad\boldsymbol{\epsilon} \sim \mathcal{N}(\boldsymbol{0}, \boldsymbol{\Sigma})
\end{align}
where $\boldsymbol{\Sigma}$ is the covariance matrix of data noise. Here, $s({\bf m}, \boldsymbol{\epsilon})$ is a known forward model that, assuming an additive noise model, can be rewritten as $F({\bf m}) + \boldsymbol{\epsilon}$, where $F({\bf m})$ represents the undersea acoustic model~\cite{jensen2011computational}. The SWellEx-96 experiment setup involves a complicated environment and no closed from analytical solution is available for $F({\bf m})$. In this case, $F({\bf m})$ can only be evaluated numerically, and we use the normal-modes program KRAKEN~\cite{porter1992kraken} for this purpose.  

Recently, machine learning algorithms have gained attention in the ocean acoustics community~\citep{BiaGerTra:19} for their notable performance and efficiency, especially when compared to traditional methods such as MCMC.
In this work, for the first time, we use the concept of invertible networks to estimate posterior distributions in GI. Particularly appealing is that invertible architectures can replace both the forward propagation model as well as the inversion method. 

We now discuss the training of the invertible architectures. 
We use the uniform prior on parameters $\mathbf{m}$ (described in $\textit{Task 1}$ and $\textit{Task 2}$ above) and the forward model in Eq.~(\ref{eq:GI-forwardmodel}) to construct a synthetic data set for the SWellEx-96 experimental setup using the normal-modes program KRAKEN. For each parameter $\mathbf{m}$, we have $7 \times 21= 147$ values for the pressure {\bf y} received at hydrophones corresponding to the source's $7$ different frequencies and the $21$ active hydrophones. For inference, we use a test acoustic signal $\mathbf{y}^*$ that corresponds to the actual parameters values from the SWellEx-96 experiment, where the source is $60$ m deep and its distance from the VLA is $3$ km (i.e. $m_1^* = 216.5$, $m_2^*=1572.368$, $m_3^*=0$, $m_4^*=23.5$). Also, the signal-to-noise ratio (SNR) is $15$ dB. 

\begin{figure}[t]
    \centering    	
    \includegraphics[scale=0.45,draft=false]{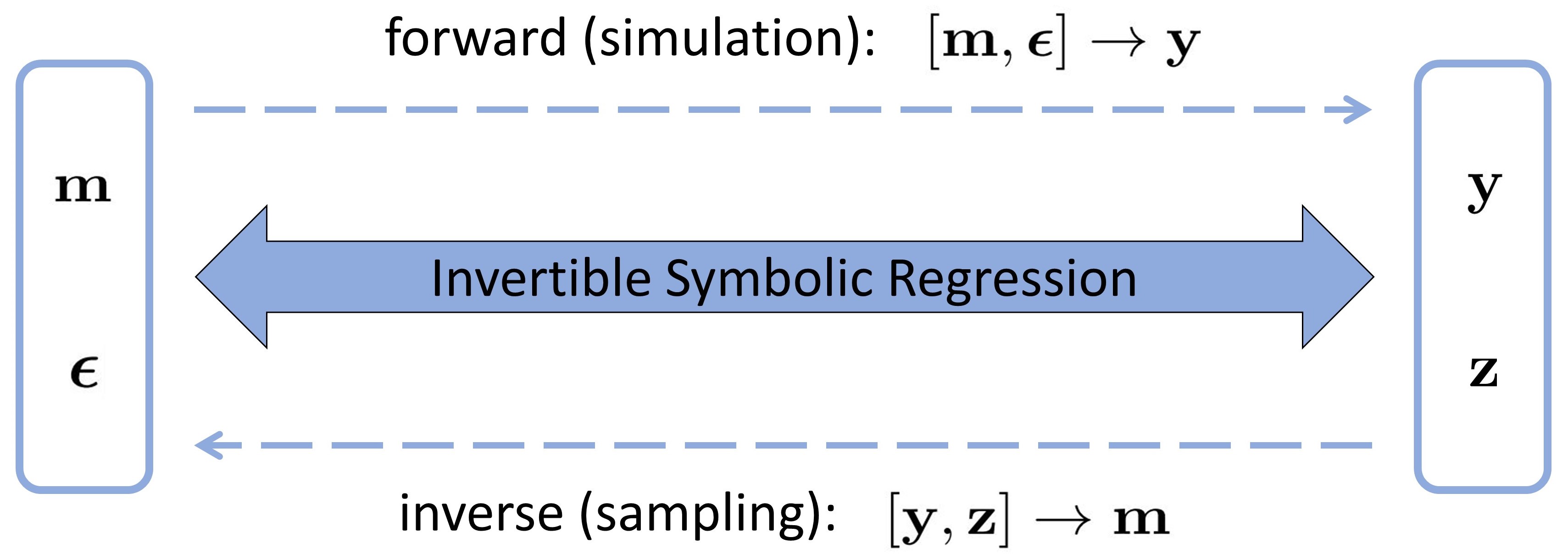}\quad\,\,
    \hspace{9pt}\includegraphics[scale=0.45,draft=false]{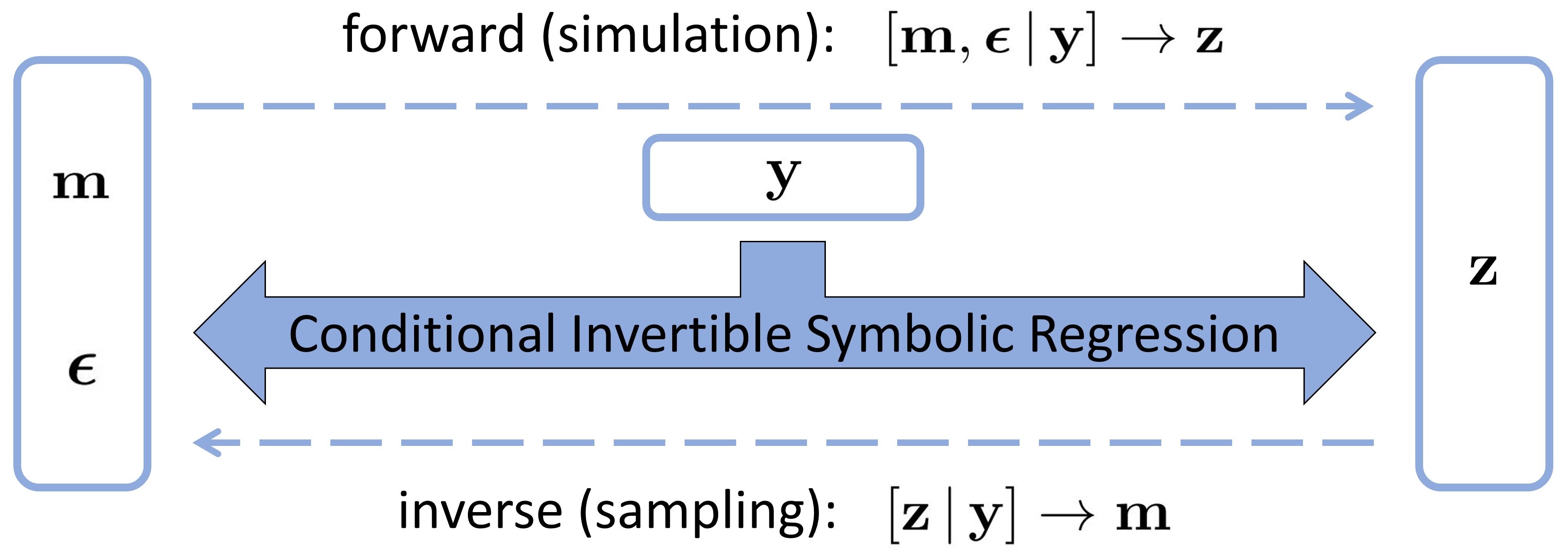}
	\caption{A conceptual figure of ISR (left) and cISR (right) for the geoacoustic inversion task. The posterior distribution of the parameters of interest $\mathbf{m}$ can be obtained by sampling $\mathbf{z}$ (e.g. from a standard Gaussian distribution) for a fixed observation $\mathbf{y}^*$ and running the trained bijective model backwards. To appropriately account for noise in the data, we include random data noise $\boldsymbol{\epsilon}$ as additional model parameters.
 }
 \label{exp_medium}
\end{figure} 

  \begin{figure}[t]
    \centering	\includegraphics[scale=0.6,draft=false]{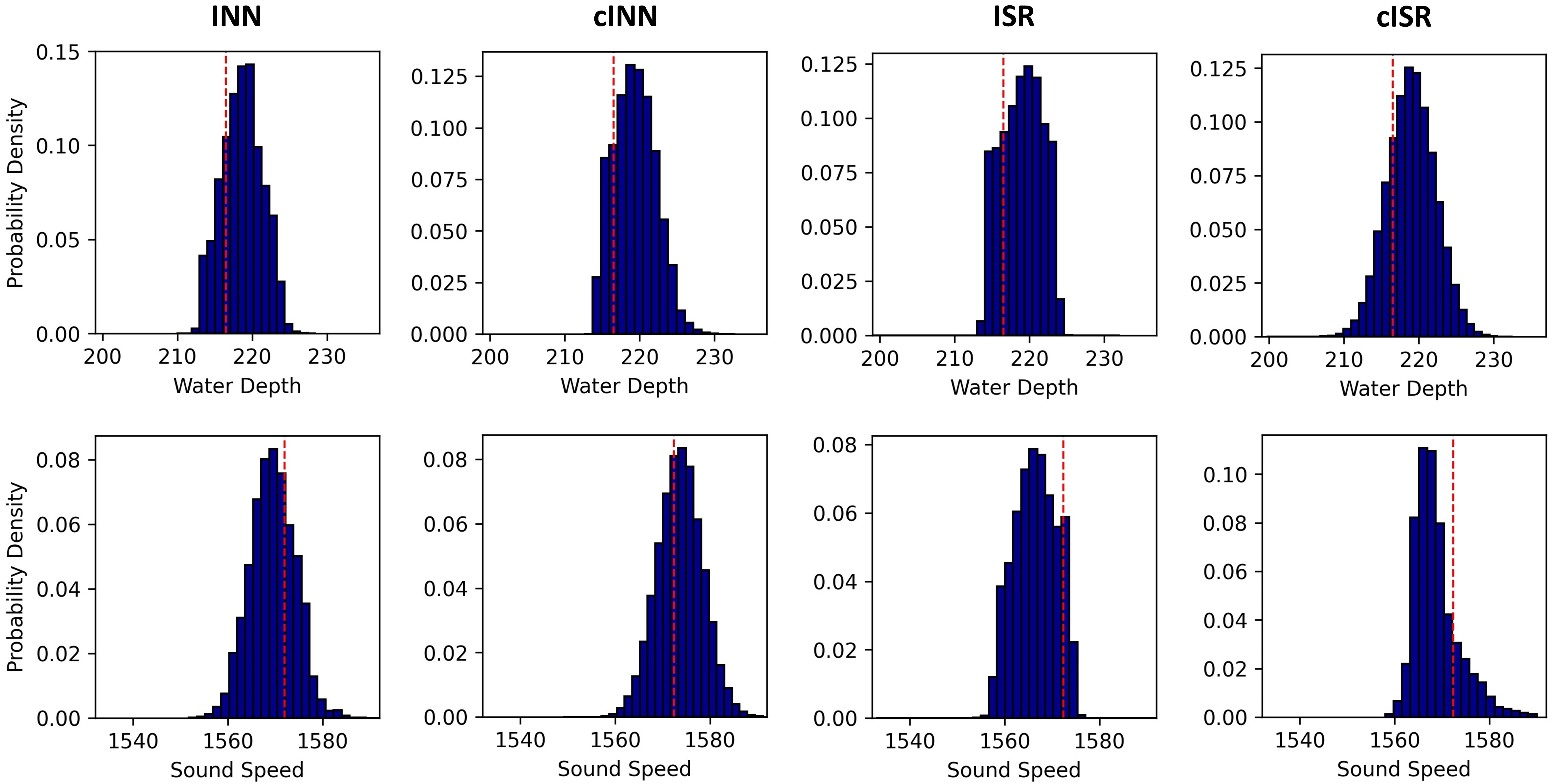}
	\caption{\textit{Task 1}. For a fixed observation $\mathbf{y}^*$, we compare the estimated posteriors $\hat{p}(\mathbf{x}\,|\,\mathbf{y}^*)$ of INN, cINN, and the proposed ISR and cISR methods. Vertical dashed red lines show the ground truth values $\mathbf{x}^*$.}
 \label{fig:GI_posteriors}
\end{figure} 

Inspired by \citet{zhang2021bayesian}, for the invertible architectures, we include data noise $\boldsymbol{\epsilon}$ as additional model parameters to be learned.
In this context, as depicted in the Figure~\ref{exp_medium}, the input of the network is obtained by augmenting the unknown parameters {\bf m} with additive noise $\boldsymbol{\epsilon}$. There are several ways to use the measurements collected across the $21$ hydrophones for training. For instance, one can stack all hydrophones' data and treat them as a single quantity at the network's output. Alternatively, one can treat each hydrophone measurement independently as an individual training example. For the former, the additive noise will be learned separately for each hydrophone pressure $\mathbf{y}$, while for the latter, we essentially learn the effective additive noise over all hydrophones simultaneously. In this experiment, we adopt the latter training approach, which disregards the inter-hydrophone variations, thereby reducing computational overhead.


\begin{figure}[t]
    \centering	
    \includegraphics[scale=0.54,draft=false]{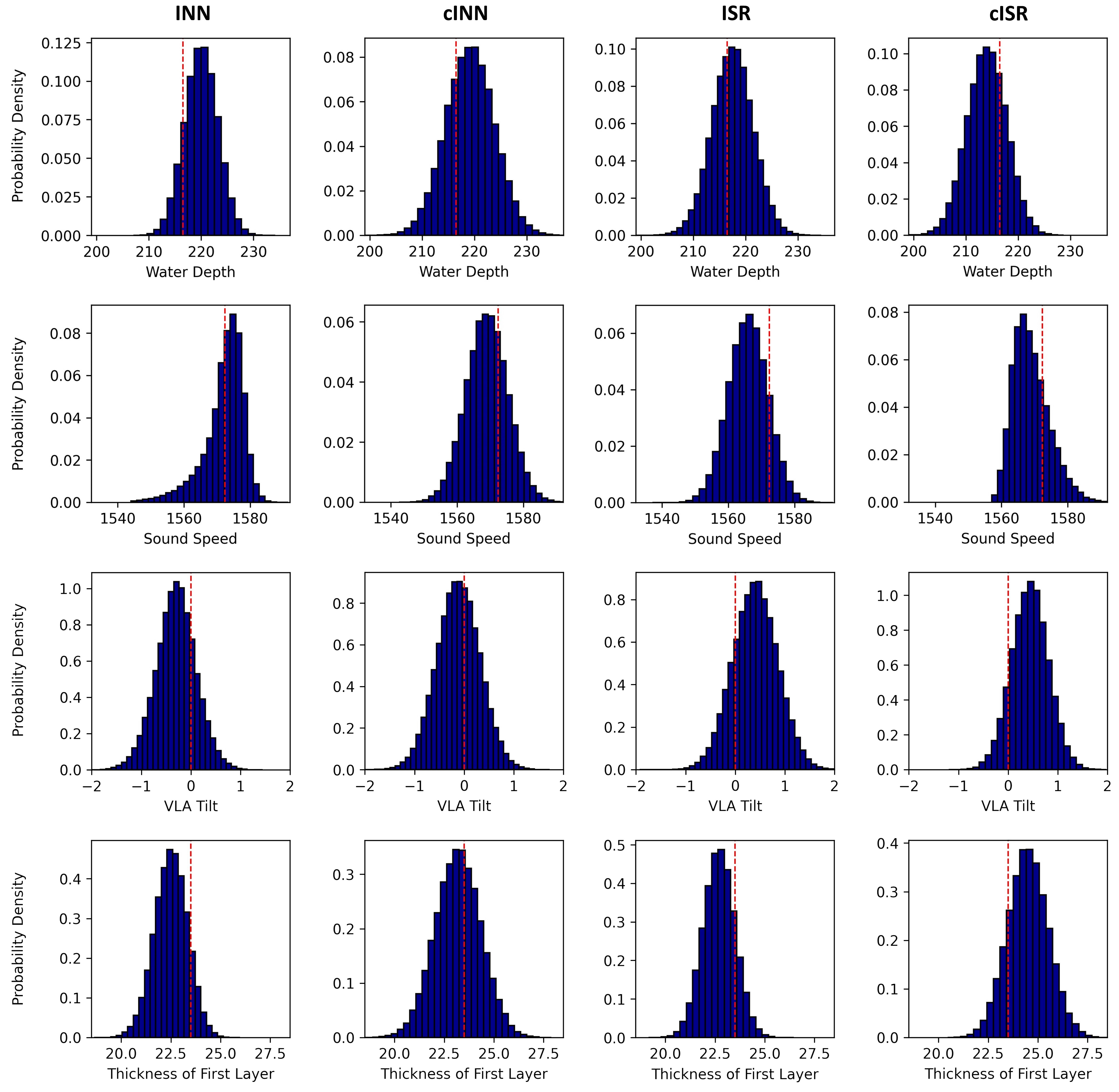}
	\caption{\textit{Task 2}. For a fixed observation $\mathbf{y}^*$, we compare the estimated posteriors $\hat{p}(\mathbf{x}\,|\,\mathbf{y}^*)$ of INN, cINN, and the proposed ISR and cISR methods. Vertical dashed red lines show the ground truth values $\mathbf{x}^*$.}
 \label{fig:GI_posteriors_4d}
\end{figure} 

The pressures received on the hydrophones are considered in the frequency domain, and hence they can be complex numbers.
While the invertible architectures can be constructed to address complex numbers, in this case study, we stack the real and imaginary parts of the pressure field at the network's output. That is, the pressure $y=\mbox{Re}\{y\}+i \, \mbox{Im} \, \{y\}$ will be represented as $\big[\mbox{Re}\{y\}, \mbox{Im} \, \{y\}\big]$ at the network's output. In short, the $7$ pressures (corresponding to the $7$ source's frequencies) received at each hydrophone are replaced by $14$ real numbers at the output of the network. 
Also, the $14$ corresponding additive noises are concatenated with the parameters $\mathbf{m}$ at the network's input. Since the dimension of the network's input is $14 + d_{\mathbf{m}}$, the latent variables $\mathbf{z}$ at the network output will be $d_{\mathbf{m}}$-dimensional for ISR and INN, and $(14 + d_{\mathbf{m}})$-dimensional for cINN and cISR. We compare the performance of the proposed ISR and cISR algorithms against INN and cINN in solving GI. 

The inferred posterior distributions via INN, cINN, ISR, and cISR, for the GI \textit{Task 1} and \textit{Task 2} are depicted in Fig.~\ref{fig:GI_posteriors} and Fig.~\ref{fig:GI_posteriors_4d},respectively. The performance of the ISR and cISR architectures is similar to that of the INN and cINN architectures. All methods produce point estimates -- Maximum a Posteriori (MAP) estimates -- close to the ground truth values, showcasing the efficacy of invertible architectures in addressing cumbersome inversion tasks.

\newpage
\section{Conclusion}
\label{sec:conclusion}
In this work, we introduce Invertible Symbolic Regression (ISR), a novel technique that identifies the relationships between the inputs and outputs of a given dataset using invertible architectures. This is achieved by bridging and integrating concepts of Invertible Neural Networks (INNs) and Equation Learner (EQL). This integration transforms the affine coupling blocks of INNs into a symbolic framework, resulting in an end-to-end differentiable symbolic inverse architecture that allows for efficient gradient-based learning. The proposed ISR method, equipped with sparsity promoting regularization, has the ability to not only capture complex functional relationships but also yield concise and interpretable invertible expressions. We demonstrate the versatility of ISR as a normalizing flow for density estimation and its applicability in solving inverse problems, particularly in the context of ocean acoustics, where it shows promising results in inferring posterior distributions of underlying parameters. This work is a first attempt toward creating interpretable symbolic invertible maps. While we mainly focused on introducing the ISR architecture and showing its applicability in density estimation tasks and inverse problems, an interesting research direction would be to explore the practicality of ISR in challenging generative modeling tasks (e.g. image or text generation, etc.). 


\subsubsection*{Acknowledgments}
T. Tohme was supported by the MathWorks Engineering Fellowship. K. Youcef-Toumi acknowledges the support from the Center for Complex Engineering Systems (CCES) at King Abdulaziz City for Science and Technology (KACST) and Massachusetts Institute of Technology (MIT). M. J. Khojasteh and F.  Meyer were supported by the Office of Naval Research under Grants  N00014-23-1-2284 and N00014-24-1-2021. \mbox{T. Tohme} also acknowledges Prof. Youssef Marzouk, Prof. Pierre Lermusiaux, and Dr. Wael H. Ali for their constructive discussions and suggestions.


\bibliography{main}

\begin{thebibliography}{93}
\providecommand{\natexlab}[1]{#1}
\providecommand{\url}[1]{\texttt{#1}}
\expandafter\ifx\csname urlstyle\endcsname\relax
  \providecommand{\doi}[1]{doi: #1}\else
  \providecommand{\doi}{doi: \begingroup \urlstyle{rm}\Url}\fi

\bibitem[Ali et~al.(2023)Ali, Charous, Mirabito, Haley, and
  Lermusiaux]{ali2023mseas}
Wael~H Ali, Aaron Charous, Chris Mirabito, Patrick~J Haley, and Pierre~FJ
  Lermusiaux.
\newblock {MSEAS-ParEq} for coupled ocean-acoustic modeling around the globe.
\newblock In \emph{OCEANS 2023-MTS/IEEE US Gulf Coast}, pp.\  1--10. IEEE,
  2023.

\bibitem[Andrieu et~al.(2003)Andrieu, De~Freitas, Doucet, and
  Jordan]{andrieu2003introduction}
Christophe Andrieu, Nando De~Freitas, Arnaud Doucet, and Michael~I Jordan.
\newblock An introduction to {MCMC} for machine learning.
\newblock \emph{Machine learning}, 50:\penalty0 5--43, 2003.

\bibitem[Ardizzone et~al.(2019{\natexlab{a}})Ardizzone, Kruse, Rother, and
  K{\"o}the]{ArdKruRotKot:19}
Lynton Ardizzone, Jakob Kruse, Carsten Rother, and Ullrich K{\"o}the.
\newblock Analyzing inverse problems with invertible neural networks.
\newblock In \emph{International Conference on Learning Representations},
  2019{\natexlab{a}}.

\bibitem[Ardizzone et~al.(2019{\natexlab{b}})Ardizzone, L{\"u}th, Kruse,
  Rother, and K{\"o}the]{ardizzone2019guided}
Lynton Ardizzone, Carsten L{\"u}th, Jakob Kruse, Carsten Rother, and Ullrich
  K{\"o}the.
\newblock Guided image generation with conditional invertible neural networks.
\newblock \emph{arXiv preprint arXiv:1907.02392}, 2019{\natexlab{b}}.

\bibitem[Ardizzone et~al.(2021)Ardizzone, Kruse, L{\"u}th, Bracher, Rother, and
  K{\"o}the]{cINN_ardizone}
Lynton Ardizzone, Jakob Kruse, Carsten L{\"u}th, Niels Bracher, Carsten Rother,
  and Ullrich K{\"o}the.
\newblock Conditional invertible neural networks for diverse image-to-image
  translation.
\newblock In Zeynep Akata, Andreas Geiger, and Torsten Sattler (eds.),
  \emph{Pattern Recognition}, pp.\  373--387, Cham, 2021. Springer
  International Publishing.
\newblock ISBN 978-3-030-71278-5.

\bibitem[Atchad{\'e} \& Rosenthal(2005)Atchad{\'e} and
  Rosenthal]{atchade2005adaptive}
Yves~F Atchad{\'e} and Jeffrey~S Rosenthal.
\newblock On adaptive {Markov chain Monte Carlo} algorithms.
\newblock \emph{Bernoulli}, 11\penalty0 (5):\penalty0 815--828, 2005.

\bibitem[Benson et~al.(2000)Benson, Chapman, and
  Antoniou]{benson2000geoacoustic}
Jeremy Benson, N~Ross Chapman, and Andreas Antoniou.
\newblock Geoacoustic model inversion using artificial neural networks.
\newblock \emph{Inverse Problems}, 16\penalty0 (6):\penalty0 1627, 2000.

\bibitem[Bianco et~al.(2019)Bianco, Gerstoft, Traer, Ozanich, Roch, Gannot, and
  Deledalle]{BiaGerTra:19}
Michael~J Bianco, Peter Gerstoft, James Traer, Emma Ozanich, Marie~A Roch,
  Sharon Gannot, and Charles-Alban Deledalle.
\newblock Machine learning in acoustics: Theory and applications.
\newblock \emph{The Journal of the Acoustical Society of America}, 146\penalty0
  (5):\penalty0 3590--3628, 2019.

\bibitem[Biggio et~al.(2021)Biggio, Bendinelli, Neitz, Lucchi, and
  Parascandolo]{biggio2021neural}
Luca Biggio, Tommaso Bendinelli, Alexander Neitz, Aurelien Lucchi, and
  Giambattista Parascandolo.
\newblock Neural symbolic regression that scales.
\newblock In \emph{International Conference on Machine Learning}, 2021.

\bibitem[Blei et~al.(2017)Blei, Kucukelbir, and McAuliffe]{blei2017variational}
David~M Blei, Alp Kucukelbir, and Jon~D McAuliffe.
\newblock Variational inference: A review for statisticians.
\newblock \emph{Journal of the American statistical Association}, 112\penalty0
  (518):\penalty0 859--877, 2017.

\bibitem[Brooks et~al.(2011)Brooks, Gelman, Jones, and
  Meng]{brooks2011handbook}
Steve Brooks, Andrew Gelman, Galin Jones, and Xiao-Li Meng.
\newblock \emph{Handbook of {Markov chain Monte Carlo}}.
\newblock CRC press, 2011.

\bibitem[Burlacu et~al.(2020)Burlacu, Kronberger, and Kommenda]{operonc++}
Bogdan Burlacu, Gabriel Kronberger, and Michael Kommenda.
\newblock Operon {C}++: An efficient genetic programming framework for symbolic
  regression.
\newblock In \emph{Proceedings of the 2020 Genetic and Evolutionary Computation
  Conference Companion}, GECCO '20, pp.\  1562–1570, New York, NY, USA, 2020.
  Association for Computing Machinery.
\newblock ISBN 9781450371278.
\newblock \doi{10.1145/3377929.3398099}.

\bibitem[Chapman \& Shang(2021)Chapman and Shang]{chapman2021review}
N~Ross Chapman and Er~Chang Shang.
\newblock Review of geoacoustic inversion in underwater acoustics.
\newblock \emph{Journal of Theoretical and Computational Acoustics},
  29\penalty0 (03):\penalty0 2130004, 2021.

\bibitem[Conrad et~al.(2016)Conrad, Marzouk, Pillai, and
  Smith]{conrad2016accelerating}
Patrick~R Conrad, Youssef~M Marzouk, Natesh~S Pillai, and Aaron Smith.
\newblock Accelerating asymptotically exact {MCMC} for computationally
  intensive models via local approximations.
\newblock \emph{Journal of the American Statistical Association}, 111\penalty0
  (516):\penalty0 1591--1607, 2016.

\bibitem[Constantine et~al.(2016)Constantine, Kent, and
  Bui-Thanh]{constantine2016accelerating}
Paul~G Constantine, Carson Kent, and Tan Bui-Thanh.
\newblock Accelerating {Markov chain Monte Carlo} with active subspaces.
\newblock \emph{SIAM Journal on Scientific Computing}, 38\penalty0
  (5):\penalty0 A2779--A2805, 2016.

\bibitem[Cranmer et~al.(2020{\natexlab{a}})Cranmer, Brehmer, and
  Louppe]{cranmer2020frontier}
Kyle Cranmer, Johann Brehmer, and Gilles Louppe.
\newblock The frontier of simulation-based inference.
\newblock \emph{Proceedings of the National Academy of Sciences}, 117\penalty0
  (48):\penalty0 30055--30062, 2020{\natexlab{a}}.

\bibitem[Cranmer et~al.(2020{\natexlab{b}})Cranmer, Sanchez~Gonzalez,
  Battaglia, Xu, Cranmer, Spergel, and Ho]{cranmer2020discovering}
Miles Cranmer, Alvaro Sanchez~Gonzalez, Peter Battaglia, Rui Xu, Kyle Cranmer,
  David Spergel, and Shirley Ho.
\newblock Discovering symbolic models from deep learning with inductive biases.
\newblock \emph{Advances in Neural Information Processing Systems},
  33:\penalty0 17429--17442, 2020{\natexlab{b}}.

\bibitem[Csill{\'e}ry et~al.(2010)Csill{\'e}ry, Blum, Gaggiotti, and
  Fran{\c{c}}ois]{csillery2010approximate}
Katalin Csill{\'e}ry, Michael~GB Blum, Oscar~E Gaggiotti, and Olivier
  Fran{\c{c}}ois.
\newblock Approximate {B}ayesian computation (abc) in practice.
\newblock \emph{Trends in ecology \& evolution}, 25\penalty0 (7):\penalty0
  410--418, 2010.

\bibitem[Dinh et~al.(2014)Dinh, Krueger, and Bengio]{dinh2014nice}
Laurent Dinh, David Krueger, and Yoshua Bengio.
\newblock Nice: Non-linear independent components estimation.
\newblock \emph{arXiv preprint arXiv:1410.8516}, 2014.

\bibitem[Dinh et~al.(2016)Dinh, Sohl-Dickstein, and Bengio]{dinh2016density}
Laurent Dinh, Jascha Sohl-Dickstein, and Samy Bengio.
\newblock Density estimation using real {NVP}.
\newblock \emph{arXiv preprint arXiv:1605.08803}, 2016.

\bibitem[Dosso \& Dettmer(2011)Dosso and Dettmer]{DosDet:11}
Stan~E Dosso and Jan Dettmer.
\newblock Bayesian matched-field geoacoustic inversion.
\newblock \emph{Inverse Problems}, 27\penalty0 (5):\penalty0 055009, 2011.

\bibitem[Doucet \& Wang(2005)Doucet and Wang]{doucet2005monte}
Arnaud Doucet and Xiaodong Wang.
\newblock {Monte Carlo} methods for signal processing: a review in the
  statistical signal processing context.
\newblock \emph{IEEE Signal Processing Magazine}, 22\penalty0 (6):\penalty0
  152--170, 2005.

\bibitem[Durkan et~al.(2019)Durkan, Bekasov, Murray, and
  Papamakarios]{durkan2019neural}
Conor Durkan, Artur Bekasov, Iain Murray, and George Papamakarios.
\newblock Neural spline flows.
\newblock \emph{Advances in neural information processing systems}, 32, 2019.

\bibitem[Edwards(1984)]{edwards1984likelihood}
Anthony William~Fairbank Edwards.
\newblock \emph{Likelihood}.
\newblock CUP Archive, 1984.

\bibitem[Fan et~al.(2014)Fan, Zurada, and Wu]{fan2014convergence}
Qinwei Fan, Jacek~M Zurada, and Wei Wu.
\newblock Convergence of online gradient method for feedforward neural networks
  with smoothing l1/2 regularization penalty.
\newblock \emph{Neurocomputing}, 131:\penalty0 208--216, 2014.

\bibitem[Gilpin et~al.(2018)Gilpin, Bau, Yuan, Bajwa, Specter, and
  Kagal]{gilpin2018explaining}
Leilani~H Gilpin, David Bau, Ben~Z Yuan, Ayesha Bajwa, Michael Specter, and
  Lalana Kagal.
\newblock Explaining explanations: An overview of interpretability of machine
  learning.
\newblock In \emph{2018 IEEE 5th International Conference on data science and
  advanced analytics (DSAA)}, pp.\  80--89. IEEE, 2018.

\bibitem[Goodfellow et~al.(2016)Goodfellow, Bengio, and
  Courville]{goodfellow2016deep}
Ian Goodfellow, Yoshua Bengio, and Aaron Courville.
\newblock \emph{Deep learning}.
\newblock MIT press, 2016.

\bibitem[Gretton et~al.(2012)Gretton, Borgwardt, Rasch, Sch{\"o}lkopf, and
  Smola]{gretton2012kernel}
Arthur Gretton, Karsten~M Borgwardt, Malte~J Rasch, Bernhard Sch{\"o}lkopf, and
  Alexander Smola.
\newblock A kernel two-sample test.
\newblock \emph{The Journal of Machine Learning Research}, 13\penalty0
  (1):\penalty0 723--773, 2012.

\bibitem[Grover et~al.(2018)Grover, Dhar, and Ermon]{grover2018flow}
Aditya Grover, Manik Dhar, and Stefano Ermon.
\newblock Flow-{GAN}: Combining maximum likelihood and adversarial learning in
  generative models.
\newblock In \emph{Proceedings of the AAAI conference on artificial
  intelligence}, volume~32, 2018.

\bibitem[Holland et~al.(2005)Holland, Dettmer, and Dosso]{holland2005remote}
Charles~W Holland, Jan Dettmer, and Stan~E Dosso.
\newblock Remote sensing of sediment density and velocity gradients in the
  transition layer.
\newblock \emph{The Journal of the Acoustical Society of America}, 118\penalty0
  (1):\penalty0 163--177, 2005.

\bibitem[Huang et~al.(2006)Huang, Gerstoft, and Hodgkiss]{huang2006uncertainty}
Chen-Fen Huang, Peter Gerstoft, and William~S Hodgkiss.
\newblock Uncertainty analysis in matched-field geoacoustic inversions.
\newblock \emph{The Journal of the Acoustical Society of America}, 119\penalty0
  (1):\penalty0 197--207, 2006.

\bibitem[Jaini et~al.(2019)Jaini, Selby, and Yu]{jaini2019sum}
Priyank Jaini, Kira~A Selby, and Yaoliang Yu.
\newblock Sum-of-squares polynomial flow.
\newblock In \emph{International Conference on Machine Learning}, pp.\
  3009--3018. PMLR, 2019.

\bibitem[Jensen et~al.(2011)Jensen, Kuperman, Porter, Schmidt, and
  Tolstoy]{jensen2011computational}
Finn~B Jensen, William~A Kuperman, Michael~B Porter, Henrik Schmidt, and
  Alexandra Tolstoy.
\newblock \emph{Computational ocean acoustics}, volume 2011.
\newblock Springer, 2011.

\bibitem[Jin et~al.(2019)Jin, Fu, Kang, Guo, and Guo]{jin2019bayesian}
Ying Jin, Weilin Fu, Jian Kang, Jiadong Guo, and Jian Guo.
\newblock Bayesian symbolic regression.
\newblock \emph{arXiv preprint arXiv:1910.08892}, 2019.

\bibitem[Kamienny et~al.(2022)Kamienny, d'Ascoli, Lample, and
  Charton]{kamienny2022end}
Pierre-Alexandre Kamienny, St{\'e}phane d'Ascoli, Guillaume Lample, and
  Fran{\c{c}}ois Charton.
\newblock End-to-end symbolic regression with transformers.
\newblock \emph{arXiv preprint arXiv:2204.10532}, 2022.

\bibitem[Keren et~al.(2023)Keren, Liberzon, and
  Lazebnik]{keren2023computational}
Liron~Simon Keren, Alex Liberzon, and Teddy Lazebnik.
\newblock A computational framework for physics-informed symbolic regression
  with straightforward integration of domain knowledge.
\newblock \emph{Scientific Reports}, 13\penalty0 (1):\penalty0 1249, 2023.

\bibitem[Kim et~al.(2020)Kim, Lu, Mukherjee, Gilbert, Jing, {\v{C}}eperi{\'c},
  and Solja{\v{c}}i{\'c}]{kim2020integration}
Samuel Kim, Peter~Y Lu, Srijon Mukherjee, Michael Gilbert, Li~Jing, Vladimir
  {\v{C}}eperi{\'c}, and Marin Solja{\v{c}}i{\'c}.
\newblock Integration of neural network-based symbolic regression in deep
  learning for scientific discovery.
\newblock \emph{IEEE Transactions on Neural Networks and Learning Systems},
  2020.

\bibitem[Kingma \& Dhariwal(2018)Kingma and Dhariwal]{kingma2018glow}
Durk~P Kingma and Prafulla Dhariwal.
\newblock Glow: Generative flow with invertible 1x1 convolutions.
\newblock \emph{Advances in neural information processing systems}, 31, 2018.

\bibitem[Kobyzev et~al.(2020)Kobyzev, Prince, and
  Brubaker]{kobyzev2020normalizing}
Ivan Kobyzev, Simon~JD Prince, and Marcus~A Brubaker.
\newblock Normalizing flows: An introduction and review of current methods.
\newblock \emph{IEEE transactions on pattern analysis and machine
  intelligence}, 43\penalty0 (11):\penalty0 3964--3979, 2020.

\bibitem[Kommenda et~al.(2020)Kommenda, Burlacu, Kronberger, and
  Affenzeller]{kommenda2020parameter}
Michael Kommenda, Bogdan Burlacu, Gabriel Kronberger, and Michael Affenzeller.
\newblock Parameter identification for symbolic regression using nonlinear
  least squares.
\newblock \emph{Genetic Programming and Evolvable Machines}, 21\penalty0
  (3):\penalty0 471--501, 2020.

\bibitem[Korattikara et~al.(2014)Korattikara, Chen, and
  Welling]{korattikara2014austerity}
Anoop Korattikara, Yutian Chen, and Max Welling.
\newblock Austerity in {MCMC} land: Cutting the metropolis-hastings budget.
\newblock In \emph{International conference on machine learning}, pp.\
  181--189. PMLR, 2014.

\bibitem[Koza \& Koza(1992)Koza and Koza]{koza1992genetic}
John~R Koza and John~R Koza.
\newblock \emph{Genetic programming: on the programming of computers by means
  of natural selection}, volume~1.
\newblock MIT press, 1992.

\bibitem[Kruse et~al.(2021)Kruse, Ardizzone, Rother, and
  K{\"o}the]{kruse2021benchmarking}
Jakob Kruse, Lynton Ardizzone, Carsten Rother, and Ullrich K{\"o}the.
\newblock Benchmarking invertible architectures on inverse problems.
\newblock \emph{arXiv preprint arXiv:2101.10763}, 2021.

\bibitem[Kungurtsev et~al.(2023)Kungurtsev, Cobb, Javidi, and
  Jalaian]{kungurtsev2023decentralized}
Vyacheslav Kungurtsev, Adam Cobb, Tara Javidi, and Brian Jalaian.
\newblock Decentralized {Bayesian} learning with {M}etropolis-adjusted
  {Hamiltonian Monte Carlo}.
\newblock \emph{Machine Learning}, pp.\  1--29, 2023.

\bibitem[La~Cava et~al.(2021)La~Cava, Orzechowski, Burlacu, de~Franca,
  Virgolin, Jin, Kommenda, and Moore]{la2021contemporary}
William La~Cava, Patryk Orzechowski, Bogdan Burlacu, Fabricio~Olivetti
  de~Franca, Marco Virgolin, Ying Jin, Michael Kommenda, and Jason~H Moore.
\newblock Contemporary symbolic regression methods and their relative
  performance.
\newblock 2021.

\bibitem[Lee(1997)]{lee1997bayesian}
Peter~M Lee.
\newblock \emph{Bayesian statistics}.
\newblock Arnold Publication, 1997.

\bibitem[Leonard \& Hsu(2001)Leonard and Hsu]{leonard2001bayesian}
Thomas Leonard and John~SJ Hsu.
\newblock \emph{Bayesian methods: an analysis for statisticians and
  interdisciplinary researchers}, volume~5.
\newblock Cambridge University Press, 2001.

\bibitem[Liu \& Tegmark(2021)Liu and Tegmark]{liu2021machine}
Ziming Liu and Max Tegmark.
\newblock Machine learning conservation laws from trajectories.
\newblock \emph{Physical Review Letters}, 126\penalty0 (18):\penalty0 180604,
  2021.

\bibitem[Liu et~al.(2024)Liu, Wang, Vaidya, Ruehle, Halverson,
  Solja{\v{c}}i{\'c}, Hou, and Tegmark]{liu2024kan}
Ziming Liu, Yixuan Wang, Sachin Vaidya, Fabian Ruehle, James Halverson, Marin
  Solja{\v{c}}i{\'c}, Thomas~Y Hou, and Max Tegmark.
\newblock Kan: Kolmogorov-arnold networks.
\newblock \emph{arXiv preprint arXiv:2404.19756}, 2024.

\bibitem[Luce et~al.(2023)Luce, Mahdavi, Wankerl, and Marquardt]{Luce_2023}
Alexander Luce, Ali Mahdavi, Heribert Wankerl, and Florian Marquardt.
\newblock Investigation of inverse design of multilayer thin-films with
  conditional invertible neural networks.
\newblock \emph{Machine Learning: Science and Technology}, 4\penalty0
  (1):\penalty0 015014, feb 2023.
\newblock \doi{10.1088/2632-2153/acb48d}.
\newblock URL \url{https://dx.doi.org/10.1088/2632-2153/acb48d}.

\bibitem[MacKay(2003)]{mackay2003information}
David~JC MacKay.
\newblock \emph{Information theory, inference and learning algorithms}.
\newblock Cambridge university press, 2003.

\bibitem[Martius \& Lampert(2016)Martius and Lampert]{martius2016extrapolation}
Georg Martius and Christoph~H Lampert.
\newblock Extrapolation and learning equations.
\newblock \emph{arXiv preprint arXiv:1610.02995}, 2016.

\bibitem[Meyer \& Gemba(2021)Meyer and Gemba]{MeyGem:J21}
Florian Meyer and Kay~L. Gemba.
\newblock Probabilistic focalization for shallow water localization.
\newblock \emph{J. Acoust. Soc. Am.}, 150\penalty0 (2):\penalty0 1057--1066, 08
  2021.

\bibitem[Mundhenk et~al.(2021)Mundhenk, Landajuela, Glatt, Santiago, Faissol,
  and Petersen]{mundhenk2021symbolic}
T~Nathan Mundhenk, Mikel Landajuela, Ruben Glatt, Claudio~P Santiago, Daniel~M
  Faissol, and Brenden~K Petersen.
\newblock Symbolic regression via neural-guided genetic programming population
  seeding.
\newblock \emph{arXiv preprint arXiv:2111.00053}, 2021.

\bibitem[Murphy(2012)]{murphy2012machine}
Kevin~P Murphy.
\newblock \emph{Machine learning: a probabilistic perspective}.
\newblock MIT press, 2012.

\bibitem[No{\'e} et~al.(2019)No{\'e}, Olsson, K{\"o}hler, and
  Wu]{noe2019boltzmann}
Frank No{\'e}, Simon Olsson, Jonas K{\"o}hler, and Hao Wu.
\newblock {Boltzmann} generators: Sampling equilibrium states of many-body
  systems with deep learning.
\newblock \emph{Science}, 365\penalty0 (6457):\penalty0 eaaw1147, 2019.

\bibitem[Orzechowski et~al.(2018)Orzechowski, La~Cava, and
  Moore]{orzechowski2018we}
Patryk Orzechowski, William La~Cava, and Jason~H Moore.
\newblock Where are we now? {A} large benchmark study of recent symbolic
  regression methods.
\newblock In \emph{Proceedings of the Genetic and Evolutionary Computation
  Conference}, pp.\  1183--1190, 2018.

\bibitem[Padmanabha \& Zabaras(2021)Padmanabha and
  Zabaras]{padmanabha2021solving}
Govinda~Anantha Padmanabha and Nicholas Zabaras.
\newblock Solving inverse problems using conditional invertible neural
  networks.
\newblock \emph{Journal of Computational Physics}, 433:\penalty0 110194, 2021.

\bibitem[Papamakarios et~al.(2019)Papamakarios, Sterratt, and
  Murray]{papamakarios2019sequential}
George Papamakarios, David Sterratt, and Iain Murray.
\newblock Sequential neural likelihood: Fast likelihood-free inference with
  autoregressive flows.
\newblock In \emph{The 22nd International Conference on Artificial Intelligence
  and Statistics}, pp.\  837--848. PMLR, 2019.

\bibitem[Papamakarios et~al.(2021)Papamakarios, Nalisnick, Rezende, Mohamed,
  and Lakshminarayanan]{papamakarios2021normalizing}
George Papamakarios, Eric Nalisnick, Danilo~Jimenez Rezende, Shakir Mohamed,
  and Balaji Lakshminarayanan.
\newblock Normalizing flows for probabilistic modeling and inference.
\newblock \emph{Journal of Machine Learning Research}, 22\penalty0
  (57):\penalty0 1--64, 2021.

\bibitem[Pawitan(2001)]{pawitan2001all}
Yudi Pawitan.
\newblock \emph{In all likelihood: statistical modelling and inference using
  likelihood}.
\newblock Oxford University Press, 2001.

\bibitem[Petersen et~al.(2021)Petersen, Larma, Mundhenk, Santiago, Kim, and
  Kim]{petersen2021deep}
Brenden~K Petersen, Mikel~Landajuela Larma, Terrell~N. Mundhenk, Claudio~Prata
  Santiago, Soo~Kyung Kim, and Joanne~Taery Kim.
\newblock Deep symbolic regression: Recovering mathematical expressions from
  data via risk-seeking policy gradients.
\newblock In \emph{International Conference on Learning Representations}, 2021.

\bibitem[Porter(1992)]{porter1992kraken}
Michael~B Porter.
\newblock The kraken normal mode program.
\newblock \emph{Naval Research Laboratory, Washington DC}, 1992.

\bibitem[Radev et~al.(2021)Radev, Graw, Chen, Mutters, Eichel,
  B{\"a}rnighausen, and K{\"o}the]{radev2021outbreakflow}
Stefan~T Radev, Frederik Graw, Simiao Chen, Nico~T Mutters, Vanessa~M Eichel,
  Till B{\"a}rnighausen, and Ullrich K{\"o}the.
\newblock Outbreakflow: Model-based {B}ayesian inference of disease outbreak
  dynamics with invertible neural networks and its application to the covid-19
  pandemics in germany.
\newblock \emph{PLoS computational biology}, 17\penalty0 (10):\penalty0
  e1009472, 2021.

\bibitem[Raissi et~al.(2019)Raissi, Perdikaris, and
  Karniadakis]{raissi2019physics}
Maziar Raissi, Paris Perdikaris, and George~E Karniadakis.
\newblock Physics-informed neural networks: A deep learning framework for
  solving forward and inverse problems involving nonlinear partial differential
  equations.
\newblock \emph{Journal of Computational physics}, 378:\penalty0 686--707,
  2019.

\bibitem[Ren et~al.(2020)Ren, Padilla, and Malof]{ren2020benchmarking}
Simiao Ren, Willie Padilla, and Jordan Malof.
\newblock Benchmarking deep inverse models over time, and the neural-adjoint
  method.
\newblock \emph{Advances in Neural Information Processing Systems},
  33:\penalty0 38--48, 2020.

\bibitem[Rezende \& Mohamed(2015)Rezende and Mohamed]{rezende2015variational}
Danilo Rezende and Shakir Mohamed.
\newblock Variational inference with normalizing flows.
\newblock In \emph{International conference on machine learning}, pp.\
  1530--1538. PMLR, 2015.

\bibitem[Sahoo et~al.(2018)Sahoo, Lampert, and Martius]{sahoo2018learning}
Subham Sahoo, Christoph Lampert, and Georg Martius.
\newblock Learning equations for extrapolation and control.
\newblock In \emph{International Conference on Machine Learning}, pp.\
  4442--4450. PMLR, 2018.

\bibitem[Salimans et~al.(2015)Salimans, Kingma, and
  Welling]{salimans2015markov}
Tim Salimans, Diederik Kingma, and Max Welling.
\newblock {Markov chain Monte Carlo} and variational inference: Bridging the
  gap.
\newblock In \emph{International conference on machine learning}, pp.\
  1218--1226. PMLR, 2015.

\bibitem[Schmidt \& Lipson(2009)Schmidt and Lipson]{schmidt2009distilling}
Michael Schmidt and Hod Lipson.
\newblock Distilling free-form natural laws from experimental data.
\newblock \emph{science}, 324\penalty0 (5923):\penalty0 81--85, 2009.

\bibitem[Sheather(2004)]{sheather2004density}
Simon~J Sheather.
\newblock Density estimation.
\newblock \emph{Statistical science}, pp.\  588--597, 2004.

\bibitem[Tibshirani(1996)]{tibshirani1996regression}
Robert Tibshirani.
\newblock Regression shrinkage and selection via the lasso.
\newblock \emph{Journal of the Royal Statistical Society: Series B
  (Methodological)}, 58\penalty0 (1):\penalty0 267--288, 1996.

\bibitem[Tohme(2020)]{tohme2020bayesian}
Tony Tohme.
\newblock \emph{The {Bayesian} validation metric: a framework for probabilistic
  model calibration and validation}.
\newblock PhD thesis, Massachusetts Institute of Technology, 2020.

\bibitem[Tohme et~al.(2020)Tohme, Vanslette, and
  Youcef-Toumi]{tohme2020generalized}
Tony Tohme, Kevin Vanslette, and Kamal Youcef-Toumi.
\newblock A generalized {Bayesian} approach to model calibration.
\newblock \emph{Reliability Engineering \& System Safety}, 204:\penalty0
  107141, 2020.

\bibitem[Tohme et~al.(2023{\natexlab{a}})Tohme, Liu, and
  Youcef-Toumi]{tohme2023gsr}
Tony Tohme, Dehong Liu, and Kamal Youcef-Toumi.
\newblock {GSR}: A generalized symbolic regression approach.
\newblock \emph{Transactions on Machine Learning Research}, 2023{\natexlab{a}}.
\newblock ISSN 2835-8856.
\newblock URL \url{https://openreview.net/forum?id=lheUXtDNvP}.

\bibitem[Tohme et~al.(2023{\natexlab{b}})Tohme, Vanslette, and
  Youcef-Toumi]{tohme2023reliable}
Tony Tohme, Kevin Vanslette, and Kamal Youcef-Toumi.
\newblock Reliable neural networks for regression uncertainty estimation.
\newblock \emph{Reliability Engineering \& System Safety}, 229:\penalty0
  108811, 2023{\natexlab{b}}.

\bibitem[Tohme et~al.(2024)Tohme, Sadr, Youcef-Toumi, and
  Hadjiconstantinou]{tohme2024messy}
Tony Tohme, Mohsen Sadr, Kamal Youcef-Toumi, and Nicolas Hadjiconstantinou.
\newblock {MESSY} {E}stimation: Maximum-entropy based stochastic and symbolic
  density estimation.
\newblock \emph{Transactions on Machine Learning Research}, 2024.
\newblock ISSN 2835-8856.
\newblock URL \url{https://openreview.net/forum?id=Y2ru0LuQeS}.

\bibitem[Tzen \& Raginsky(2019)Tzen and Raginsky]{tzen2019neural}
Belinda Tzen and Maxim Raginsky.
\newblock Neural stochastic differential equations: Deep latent {Gaussian}
  models in the diffusion limit.
\newblock \emph{arXiv preprint arXiv:1905.09883}, 2019.

\bibitem[Udrescu \& Tegmark(2020)Udrescu and Tegmark]{udrescu2020aifeynman}
Silviu-Marian Udrescu and Max Tegmark.
\newblock {AI Feynman}: A physics-inspired method for symbolic regression.
\newblock \emph{Science Advances}, 6\penalty0 (16):\penalty0 eaay2631, 2020.

\bibitem[Udrescu et~al.(2020)Udrescu, Tan, Feng, Neto, Wu, and
  Tegmark]{udrescu2020aifeynman_2}
Silviu-Marian Udrescu, Andrew Tan, Jiahai Feng, Orisvaldo Neto, Tailin Wu, and
  Max Tegmark.
\newblock {AI Feynman} 2.0: {Pareto-optimal} symbolic regression exploiting
  graph modularity.
\newblock \emph{Advances in Neural Information Processing Systems},
  33:\penalty0 4860--4871, 2020.

\bibitem[Valipour et~al.(2021)Valipour, You, Panju, and
  Ghodsi]{valipour2021symbolicgpt}
Mojtaba Valipour, Bowen You, Maysum Panju, and Ali Ghodsi.
\newblock Symbolicgpt: A generative transformer model for symbolic regression.
\newblock \emph{arXiv preprint arXiv:2106.14131}, 2021.

\bibitem[Vanslette et~al.(2020)Vanslette, Tohme, and
  Youcef-Toumi]{vanslette2020general}
Kevin Vanslette, Tony Tohme, and Kamal Youcef-Toumi.
\newblock A general model validation and testing tool.
\newblock \emph{Reliability Engineering \& System Safety}, 195:\penalty0
  106684, 2020.

\bibitem[Virgolin \& Pissis(2022)Virgolin and Pissis]{virgolin2022symbolic}
Marco Virgolin and Solon~P Pissis.
\newblock Symbolic regression is {NP}-hard.
\newblock \emph{arXiv preprint arXiv:2207.01018}, 2022.

\bibitem[Wang \& Marzouk(2022)Wang and Marzouk]{wang2022minimax}
Sven Wang and Youssef Marzouk.
\newblock On minimax density estimation via measure transport.
\newblock \emph{arXiv preprint arXiv:2207.10231}, 2022.

\bibitem[Wenliang et~al.(2019)Wenliang, Sutherland, Strathmann, and
  Gretton]{wenliang2019learning}
Li~Wenliang, Danica~J Sutherland, Heiko Strathmann, and Arthur Gretton.
\newblock Learning deep kernels for exponential family densities.
\newblock In \emph{International Conference on Machine Learning}, pp.\
  6737--6746. PMLR, 2019.

\bibitem[Wild \& Seber(1989)Wild and Seber]{wild1989nonlinear}
CJ~Wild and GAF Seber.
\newblock \emph{Nonlinear regression}.
\newblock New York: Wiley, 1989.

\bibitem[Wu et~al.(2018)Wu, Nowozin, Meeds, Turner, Hern{\'a}ndez-Lobato, and
  Gaunt]{wu2018deterministic}
Anqi Wu, Sebastian Nowozin, Edward Meeds, Richard~E Turner, Jos{\'e}~Miguel
  Hern{\'a}ndez-Lobato, and Alexander~L Gaunt.
\newblock Deterministic variational inference for robust bayesian neural
  networks.
\newblock In \emph{International Conference on Learning Representations}, 2018.

\bibitem[Wu et~al.(2023)Wu, Huang, and Zhao]{WuHuaZha:23}
Sihong Wu, Qinghua Huang, and Li~Zhao.
\newblock Fast {B}ayesian inversion of airborne electromagnetic data based on
  the invertible neural network.
\newblock \emph{IEEE Transactions on Geoscience and Remote Sensing},
  61:\penalty0 1--11, 2023.

\bibitem[Wu et~al.(2014)Wu, Fan, Zurada, Wang, Yang, and Liu]{wu2014batch}
Wei Wu, Qinwei Fan, Jacek~M Zurada, Jian Wang, Dakun Yang, and Yan Liu.
\newblock Batch gradient method with smoothing l1/2 regularization for training
  of feedforward neural networks.
\newblock \emph{Neural Networks}, 50:\penalty0 72--78, 2014.

\bibitem[Xu et~al.(2010)Xu, Zhang, Wang, Chang, and Liang]{xu20101}
Zongben Xu, Hai Zhang, Yao Wang, XiangYu Chang, and Yong Liang.
\newblock L 1/2 regularization.
\newblock \emph{Science China Information Sciences}, 53:\penalty0 1159--1169,
  2010.

\bibitem[Yardim et~al.(2010)Yardim, Gerstoft, and
  Hodgkiss]{yardim2010geoacoustic}
Caglar Yardim, Peter Gerstoft, and William~S Hodgkiss.
\newblock Geoacoustic and source tracking using particle filtering:
  Experimental results.
\newblock \emph{The Journal of the Acoustical Society of America}, 128\penalty0
  (1):\penalty0 75--87, 2010.

\bibitem[Zhang et~al.(2022)Zhang, Zhou, Qian, and Zhang]{zhang2022ps}
Hengzhe Zhang, Aimin Zhou, Hong Qian, and Hu~Zhang.
\newblock {PS}-{Tree}: A piecewise symbolic regression tree.
\newblock \emph{Swarm and Evolutionary Computation}, 71:\penalty0 101061, 2022.

\bibitem[Zhang \& Curtis(2021)Zhang and Curtis]{zhang2021bayesian}
Xin Zhang and Andrew Curtis.
\newblock Bayesian geophysical inversion using invertible neural networks.
\newblock \emph{Journal of Geophysical Research: Solid Earth}, 126\penalty0
  (7):\penalty0 e2021JB022320, 2021.

\end{thebibliography}
\bibliographystyle{tmlr}

\newpage
\appendix

\counterwithin{equation}{section}
\section{Invertible symbolic expressions recovered by ISR for the considered distributions}
\label{app:a}

\begin{table}[h]
\caption{Invertible symbolic expressions recovered by our ISR method for density estimation of several distributions (see Figure \ref{fig:normalizing_flow}) in Section \ref{sec:density_estimation_normalizing_flow}. Here, MoGs denotes ``Mixture of Gaussians.'' \label{table:isr_expr}}
\vspace{-1mm}
\tabcolsep=0.2cm
\centering
\scalebox{0.6}{
\begin{tabular}{l|l}
\toprule
\textbf{Example} & \multicolumn{1}{ c }{\textbf{Expression}} \\
\midrule
Gaussian &  $\mathbf{u} = \mathbf{x}$, i.e. $u_1 = x_1$, $u_2 = x_2$\\
& $s_1(u_2) = 1.16$ \\
& $t_1(u_2) = 0$\\
& $v_1 = u_1 \cdot \exp\scalebox{1.1}{$($}s_1(u_2)\scalebox{1.1}{$)$} + t_1(u_2)$\\
& $v_2 = u_2$\\
& $s_2(v_1) = 1.14$ \\
& $t_2(v_1) = - 9.39$\\
& $o_1 = v_1$\\
& $o_2 = v_2 \cdot \exp\scalebox{1.1}{$($}s_2(v_1)\scalebox{1.1}{$)$} + t_2(v_1)$\\
&  $\mathbf{z} = \mathbf{o}$, i.e. $z_1 = o_1$, $z_2 = o_2$\\
\midrule
Banana &  $\mathbf{u} = \mathbf{x}$, i.e. $u_1 = x_1$, $u_2 = x_2$\\
& $s_1(u_2) = 0.52 \sin{\left(1.86\, u_2 \right)} + 0.084 \sin{\left(5.5 \sin{\left(1.86\, u_2 \right)} + 2.55 \right)} + 1.7$ \\
& $t_1(u_2) = 0.74 - 0.12 \sin{\left(3.65 \sin{\left(2.45\,u_2 \right)} - 0.82 \right)}$\\
& $v_1 = u_1 \cdot \exp\scalebox{1.1}{$($}s_1(u_2)\scalebox{1.1}{$)$} + t_1(u_2)$\\
& $v_2 = u_2$\\
& $s_2(v_1) = 1.72 \left(0.025 - 0.47 \sin{\left(0.33\,v_1 \right)}\right) \left(0.23 \sin{\left(0.33\,v_1 \right)} - 0.29\right) + 2.24$ \\
& $t_2(v_1) = - 3.74 \left(0.022 \sin{\left(0.62\,v_1 \right)} + 0.035 \sin{\left(0.63 \,v_1 \right)} - 0.76\right) \left(0.45 \sin{\left(0.62\,v_1 \right)} + 0.7 \sin{\left(0.63\,v_1 \right)} + 0.38\right) + 0.027$\\
& $o_1 = v_1$\\
& $o_2 = v_2 \cdot \exp\scalebox{1.1}{$($}s_2(v_1)\scalebox{1.1}{$)$} + t_2(v_1)$\\
&  $\mathbf{z} = \mathbf{o}$, i.e. $z_1 = o_1$, $z_2 = o_2$\\
\midrule

Ring &  $\mathbf{u} = \mathbf{x}$, i.e. $u_1 = x_1$, $u_2 = x_2$\\
& $s_1(u_2) = - 3.14 \left(- 0.15\,u_2^{2} - 0.26 \sin{\left(1.26\,u_2 \right)} + 0.094 \sin{\left(3.25\,u_2 \right)} - 0.3\right) \left(0.098\,u_2^{2} + 0.39 \sin{\left(1.26\,u_2 \right)} + 0.014 \sin{\left(3.25\,u_2 \right)} + 0.16\right)$\\
&\hspace{1.43cm}$ + 0.09 \sin{\left(0.17\,u_2^{2} + 0.69 \sin{\left(1.26\,u_2 \right)} + 0.76 \sin{\left(3.25\,u_2 \right)} + 0.25 \right)} - 0.30 \sin{\left(0.94\,u_2^{2} + 2.35 \sin{\left(1.26\,u_2 \right)} - 1.31 \sin{\left(3.25\,u_2 \right)} + 1.57 \right)} + 0.053$ \\

& $t_1(u_2) = -0.012$\\

& $v_1 = u_1 \cdot \exp\scalebox{1.1}{$($}s_1(u_2)\scalebox{1.1}{$)$} + t_1(u_2)$\\
& $v_2 = u_2$\\

& $s_2(v_1) = 0.037 \sin{\left(0.22 \sin{\left(0.22\,v_1 \right)} + 0.22 \sin{\left(0.22\,v_1 \right)} - 0.27 \right)} + 0.052 \sin{\left(0.23 \sin{\left(0.22\,v_1 \right)} + 0.23 \sin{\left(0.22\,v_1 \right)} - 0.39 \right)} - 0.11$ \\

& $t_2(v_1) = 0.13 \sin{\left(0.13 \sin{\left(0.37\,v_1 \right)} + 0.13 \sin{\left(0.59\,v_1 \right)} - 1.16 \right)} + 0.65 \sin{\left(0.021\,v_1 + 1.34 \sin{\left(0.37\,v_1 \right)} + 2.29 \sin{\left(0.59\,v_1 \right)} + 1.44 \right)} + 0.33$\\

& $o_1 = v_1$\\
& $o_2 = v_2 \cdot \exp\scalebox{1.1}{$($}s_2(v_1)\scalebox{1.1}{$)$} + t_2(v_1)$\\
&  $\mathbf{u} = \mathbf{o}$, i.e. $u_1 = o_1$, $u_2 = o_2$\\

& $s_1(u_2) = - 3.65 \left(- 0.47 \sin{\left(1.38\,u_2 \right)} - 0.026 \sin{\left(1.93\,u_2 \right)} - 0.1\right) \left(- 0.014 \sin{\left(1.38\,u_2 \right)} - 0.39 \sin{\left(1.93\,u_2 \right)} - 0.35\right) - 0.44 \sin{\left(1.74 \sin{\left(1.38\,u_2 \right)} + 1.71 \sin{\left(1.93\,u_2 \right)} + 5.55 \right)}$\\
&\hspace{1.31cm} $- 0.46 \sin{\left(3.31 \sin{\left(1.38\,u_2 \right)} + 0.44 \sin{\left(1.93\,u_2 \right)} - 0.7 \right)} + 0.38$ \\

& $t_1(u_2) = 0.11 \sin{\left(0.85 \sin{\left(1.38\,u_2 \right)} + 0.86 \sin{\left(1.38\,u_2 \right)} \right)} + 0.12 \sin{\left(0.87 \sin{\left(1.38\,u_2 \right)} + 0.87 \sin{\left(1.38\,u_2 \right)} \right)} + 0.053$\\

& $v_1 = u_1 \cdot \exp\scalebox{1.1}{$($}s_1(u_2)\scalebox{1.1}{$)$} + t_1(u_2)$\\
& $v_2 = u_2$\\

& $s_2(v_1) = 0.25\,v_1^{2} - 0.0071 \sin{\left(1.67\,v_1 \right)} - 0.1 \sin{\left(5.29\,v_1 \right)} - 0.63 \sin{\left(- 1.075\,v_1^{2} + 0.55 \sin{\left(1.67\,v_1 \right)} + 0.77 \sin{\left(5.29\,v_1 \right)} \right)}$\\
&\hspace{1.36cm}$ + 0.46 \sin{\left(1.89\,v_1^{2} - 0.27 \sin{\left(1.67\,v_1 \right)} - 1.69 \sin{\left(5.29\,v_1 \right)} + 0.9 \right)} + 1.22$\\

& $t_2(v_1) = 0.62 \sin{\left(1.56 \sin{\left(0.43\,v_1 \right)} + 1.57 \sin{\left(0.43\,v_1 \right)} - 1.68 \right)} + 0.61 \sin{\left(1.56 \sin{\left(0.43\,v_1 \right)} + 1.57 \sin{\left(0.43\,v_1 \right)} - 1.68 \right)} - 0.48$\\

& $o_1 = v_1$\\
& $o_2 = v_2 \cdot \exp\scalebox{1.1}{$($}s_2(v_1)\scalebox{1.1}{$)$} + t_2(v_1)$\\
&  $\mathbf{z} = \mathbf{o}$, i.e. $z_1 = o_1$, $z_2 = o_2$\\
\midrule

MoG &  $\mathbf{u} = \mathbf{x}$, i.e. $u_1 = x_1$, $u_2 = x_2$\\
& $s_1(u_2) = - 0.039 \left(- 0.032 \sin{\left(1.44\,u_2 \right)} - \sin{\left(1.48\,u_2 \right)} + 0.94\right)^{2} - 3.03\left(0.18 \sin{\left(1.44\,u_2 \right)} + 0.14 \sin{\left(1.48\,u_2 \right)} - 0.63\right) \left(0.26 \sin{\left(1.44\,u_2 \right)} + 0.28 \sin{\left(1.48\,u_2 \right)} - 0.31\right)$\\
& \hspace{1.43cm}$+ 0.047 \sin{\left(1.44\,u_2 \right)} + 0.13 \sin{\left(1.48\,u_2 \right)} - 0.029 \sin{\left(0.15 \sin{\left(1.44\,u_2 \right)} + 1.4 \sin{\left(1.48\,u_2 \right)} - 3.47 \right)} - 0.11 \sin{\left(0.45 \sin{\left(1.44\,u_2 \right)} + 1.46 \sin{\left(1.48\,u_2 \right)} + 2.65 \right)} - 0.17$ \\
& $t_1(u_2) = 0.052 - 0.12 \sin{\left(1.13 \sin{\left(3.14\,u_2 \right)} + 1.64 \right)}$\\
& $v_1 = u_1 \cdot \exp\scalebox{1.1}{$($}s_1(u_2)\scalebox{1.1}{$)$} + t_1(u_2)$\\
& $v_2 = u_2$\\
& $s_2(v_1) = 0.34 \left(0.15 \sin{\left(1.024\,v_1 \right)} + 0.13 \sin{\left(2.022\,v_1 \right)}\right) \sin{\left(2.022\,v_1 \right)} + 0.014 \sin^{2}{\left(2.022\,v_1 \right)} + 0.15 \sin{\left(0.98\,v_1 \sin{\left(1.024\,v_1 \right)} + 1.9 \sin{\left(2.022\,v_1 \right)} - 1.41 \right)}$\\
& \hspace{1.37cm}$ - 0.22 \sin{\left(3.38 \sin{\left(1.024\,v_1 \right)} + 1.83 \sin{\left(2.022\,v_1 \right)} + 1.5 \right)} + 0.39$ \\
& $t_2(v_1) = - 0.46 \sin{\left(0.89\,v_1 + 0.21 \sin{\left(1.38\,v_1 \right)} - 1.56 \right)} + 0.33 \sin{\left(1.23\,v_1 + 1.69 \sin{\left(1.38\,v_1 \right)} - 0.44 \sin{\left(1.58\,v_1 \right)} + 1.75\,v_1 \right)} + 0.094$\\
& $o_1 = v_1$\\
& $o_2 = v_2 \cdot \exp\scalebox{1.1}{$($}s_2(v_1)\scalebox{1.1}{$)$} + t_2(v_1)$\\
&  $\mathbf{u} = \mathbf{o}$, i.e. $u_1 = o_1$, $u_2 = o_2$\\
& $s_1(u_2) = 3.36 \left(- 0.36 \sin{\left(3.1\,u_2 \right)} - 0.29\right) \left(0.26 \sin{\left(3.1\,u_2 \right)} + 0.19\right) - 0.45 \sin{\left(3.88 \sin{\left(3.1\,u_2 \right)} - 1.83 \right)} - 0.38$\\
& $t_1(u_2) = 0$\\
& $v_1 = u_1 \cdot \exp\scalebox{1.1}{$($}s_1(u_2)\scalebox{1.1}{$)$} + t_1(u_2)$\\
& $v_2 = u_2$\\
& $s_2(v_1) = 0.0035\,v_1^{2} - 2.88 \left(- 0.079\,v_1^{2} - 0.33\right) \left(0.042\,v_1^{2} + 0.4\right) + 2.6 \left(- 0.34 \sin{\left(1.7\,v_1 \right)} - 0.088 \sin{\left(1.71\,v_1 \right)}\right) \left(- 0.053 \sin{\left(1.7\,v_1 \right)} - 0.38 \sin{\left(1.71\,v_1 \right)}\right) + 1.61$\\
& $t_2(v_1) = -0.14$\\
& $o_1 = v_1$\\
& $o_2 = v_2 \cdot \exp\scalebox{1.1}{$($}s_2(v_1)\scalebox{1.1}{$)$} + t_2(v_1)$\\
&  $\mathbf{z} = \mathbf{o}$, i.e. $z_1 = o_1$, $z_2 = o_2$\\
\bottomrule
\end{tabular}
}
\end{table}

\newpage
\section{Details of network architectures}
\label{app:b}

In our experiments, we train all models using Adam optimizer with a dynamic learning rate decaying from $10^{-2}$ to $10^{-4}$.
In addition, for INN, all (hidden) neurons of the subnetworks are followed by Leaky ReLU activations, while for ISR, each neuron in the hidden layers is followed by an activation function from the following library: 
\begin{align}
\big\{1, \text{id}, \textcolor{gray}{\bullet}^2(\times 4), \sin(2\pi \textcolor{gray}{\bullet}), \sigmoid, \textcolor{gray}{\bullet}_1\times\textcolor{gray}{\bullet}_2\big\}\notag
\end{align}
where $1$ represents the constant function, ``id'' is the identity operator, $\sigmoid$ denotes the sigmoid function, and ${\color{gray} \bullet}$ denotes placeholder operands, e.g. ${\color{gray} \bullet}^2$ corresponds to the square operator. Also, ${\color{gray} \bullet}_1\times{\color{gray} \bullet}_2$ denotes the multiplication operator, and each activation function may be duplicated within each layer.

The architecture details are provided below.

\textit{Density estimation via normalizing flow.} In Section \ref{sec:density_estimation_normalizing_flow}, we train INN and ISR using a batch \mbox{size of $64$.} For the ``Gaussian'' and ``Banana'' distributions, we adopt $1$ affine coupling block with $2$ fully connected (hidden) layers per subnetwork. For the ``Ring'' and ``Mixture of Gaussians (MoGs)'' distributions, we use $2$ invertible blocks with $2$ fully connected layers for each subnetwork.

\textit{Inverse Kinematics.} In Section \ref{sec:inverse_kinematics}, we train all models using a batch size of $100$. For all models, we adopt $6$ reversible blocks with $3$ fully connected layers per subnetwork. 

\textit{Geoacoustic Inversion.} In Section \ref{sec:geoacoustic_inversion}, we train all models using a batch size of $200$. For all models, we adopt $5$ invertible blocks with $4$ fully connected layers for each subnetwork.

\section{Quantitative Evaluation -- Inverse Kinematics}
\label{app:c}
Here, we quantitatively evaluate the quality of the estimated posteriors by the different models considered in the inverse kinematics experiment. To ensure a fair comparison among all methods, we use the same training data and train all models for the same number of epochs, and using identical batches and architectures (as provided in the previous section). 

As suggested in \citep{kruse2021benchmarking}, we evaluate the correctness of the estimated posteriors using two metrics.
First, we use the Maximum Mean Discrepancy (MMD) introduced by \citet{gretton2012kernel}, which computes the \emph{posterior mismatch} between the distribution $\hat{p}(\mathbf{x}\,|\,\mathbf{y}^*)$ produced by 
a model and a ground truth estimate $p_{\text{gt}}(\mathbf{x}\,|\,\mathbf{y}^*)$, which in this case is obtained via rejection sampling (see Section \ref{sec:inverse_kinematics}), i.e. 
\begin{align}
\text{Err}_{\text{post}} = \text{MMD}\big(\hat{p}(\mathbf{x}\,|\,\mathbf{y}^*), p_{\text{gt}}(\mathbf{x}\,|\,\mathbf{y}^*)\big)\label{eq:mmd_metric}
\end{align}
\vspace{-0.4cm}\\
Second, we measure the \emph{re-simulation error}, which applies the true forward process $f$ in Eq.~(\ref{eq:inverse_kinematics_forward}) to the generated samples $\mathbf{x}$ and computes the mean squared distance to the target $\mathbf{y}^*$, i.e.
\begin{align}
\text{Err}_{\text{resim}} = \mathbb{E}_{\mathbf{x}\sim\hat{p}(\mathbf{x}\,|\,\mathbf{y}^*)}\hspace{-1pt}\left[||f(\mathbf{x}) - \mathbf{y}^*||_2^2\right]\label{eq:resim_metric}
\end{align}
\vspace{-1.1cm}\\
\begin{table}[h]
\caption{Quantitative results for the inverse kinematics benchmark experiment. 
\label{table:inv_kin}}
\vspace{-1mm}
\tabcolsep=0.2cm
\centering
\begin{tabular}{l|cc}
\toprule
\textbf{Method} & $\text{Err}_{\text{post}}$ & $\text{Err}_{\text{resim}}$ \\
\midrule 
INN & 0.0259  & 0.0163\\
cINN & 0.0162 & 0.0087\\
ISR & 0.0286  & 0.0196\\
cISR & 0.0221 & 0.0134\\
\bottomrule
\end{tabular}
\end{table}

\end{document}